\documentclass[11pt, a4paper, logo, copyright, nonumbering]{klear}
\usepackage[authoryear, compress, round]{natbib}
\usepackage{dblfloatfix}
\usepackage{caption}
\usepackage{dramatist}
\usepackage{xspace}
\usepackage{pifont} 
\usepackage{multirow}
\usepackage{tcolorbox}
\usepackage{xltabular}
\usepackage{longtable}
\usepackage{hyperref}
\usepackage{amsfonts}
\usepackage{amsmath}
\usepackage{amssymb}
\usepackage{lineno}
\usepackage{multirow}
\usepackage{adjustbox}

\usepackage[bottom]{footmisc}

\usepackage{CJKutf8}
\usepackage{subfigure}
\usepackage{setspace}

\usepackage{dsfont}
\usepackage{array} 
\usepackage{tabularx} 
\usepackage{subfigure} 
\usepackage{xcolor} 
\usepackage{algorithm}
\usepackage{algorithmic}

\usepackage{lipsum}  
\usepackage{multicol} 

\usepackage{graphicx}
\usepackage{booktabs}
\usepackage{multirow}

\definecolor{keywordcolor}{rgb}{0.7, 0.1, 0.1}   
\definecolor{tacticcolor}{rgb}{0.0, 0.1, 0.6}    
\definecolor{commentcolor}{rgb}{0.4, 0.4, 0.4}   
\definecolor{symbolcolor}{rgb}{0.0, 0.1, 0.6}    
\definecolor{sortcolor}{rgb}{0.1, 0.5, 0.1}      
\definecolor{attributecolor}{rgb}{0.7, 0.1, 0.1} 

\usepackage{listings}

\usepackage{textcomp}

\usepackage{tcolorbox}
\newtcolorbox{promptbox}[1][]{
    colback=gray!10,      
    colframe=black!50,     
    boxrule=0.5mm,        
    arc=1mm,              
    boxsep=0mm,
    fontupper=\ttfamily\scriptsize,  
    width=\textwidth,     
    title=#1,
fonttitle=\ttfamily\footnotesize\centering,
}

\makeatletter
\def\@BTrule[#1]{%
  \ifx\longtable\undefined
    \let\@BTswitch\@BTnormal
  \else\ifx\hline\LT@hline
    \nobreak
    \let\@BTswitch\@BLTrule
  \else
     \let\@BTswitch\@BTnormal
  \fi\fi
  \global\@thisrulewidth=#1\relax
  \ifnum\@thisruleclass=\tw@\vskip\@aboverulesep\else
  \ifnum\@lastruleclass=\z@\vskip\@aboverulesep\else
  \ifnum\@lastruleclass=\@ne\vskip\doublerulesep\fi\fi\fi
  \@BTswitch}
\makeatother

\addto\extrasenglish{
}

 {\begin{list}{}%
         {\setlength{\leftmargin}{#1}}%
         \item[]%
 }
 {\end{list}}
 
\reportnumber{001} 

\renewcommand{\today}{}

\newcommand{\kimina}{~Kimina-Prover-distilled-7B}
\newcommand{\dsii}{~DeepSeek-Prover-V2-distilled-7B}
\newcommand{\leani}{~\texttt{Leanabell-Prover-V1}}
\newcommand{\leanii}{~\texttt{Leanabell-Prover-V2}}
\newcommand{\kylean}{Leanabell-Prover}

\newcommand{\goedelii}{Goedel-Prover-SFT}
\newcommand{\dsi}{DeepSeek-Prover-v1.5}

{\centering
\title{Leanabell-Prover-V2: Verifier-integrated Reasoning for Formal Theorem Proving via Reinforcement Learning}}

\author[*]{
\quad Xingguang Ji, Yahui Liu, Qi Wang$^{\heartsuit}$, Jingyuan Zhang, Yang Yue, Rui Shi, Chenxi Sun, \newline Fuzheng Zhang,  Guorui Zhou, Kun Gai
\\
Klear Team, Kuaishou Technology
}

\correspondingauthor{Corresponding author.}

\begin{abstract}

We introduce our \leanii, a 7B large language models (LLMs) that can produce formal theorem proofs in Lean 4, with verifier-integrated Long Chain-of-Thoughts (CoT).
Following our previous work \leani, we continual to choose to posttrain existing strong prover models for further performance improvement. In our V2 version, we mainly upgrade the Reinforcement Learning (RL) with feedback provided by the Lean 4 verifier.
Crucially, verifier feedback, such as indicating success or detailing specific errors, allows the LLM to become ``self-aware'' of the correctness of its own reasoning process and learn to reflexively correct errors. \leanii~ directly optimizes LLM reasoning trajectories with multi-turn verifier interactions, together with feedback token masking for stable RL training and a simple reward strategy. %
Experiments show that \leanii~ improves performance by 3.2\% (pass@128) with Kimina-Prover-Preview-Distill-7B and 2.0\% (pass@128) with DeepSeek-Prover-V2-7B on the MiniF2F test set. The source codes, curated data and models are available at: \url{https://github.com/Leanabell-LM/Leanabell-Prover-V2}.
\end{abstract}

\begin{document}
\begin{CJK*}{UTF8}{gbsn}

\maketitle

\begin{figure}[h]
\centering
\includegraphics[width=0.65\linewidth]{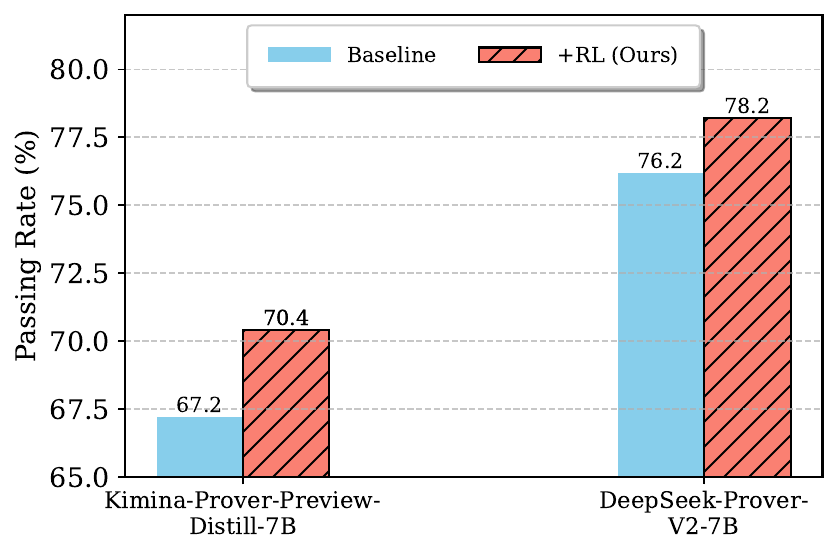}
\caption{
    Benchmark performance on MiniF2F-test~\citep{zheng2021minif2f}. Our method boosts both the two baseline models after employing RL training.  Our framework surpasses Kimina-Prover-Preview-Distill-7B and DeepSeek-Prover-V2-7B 3.2\% and 2.0\% at pass@128, respectively.
}
\label{fig:RL_performance}
\end{figure}

\newpage

\section{Introduction}
\label{sec:introduction}

Formal reasoning offers powerful techniques for verifying the correctness of complex mathematical theorems, providing a level of assurance unattainable through traditional testing alone~\citep{huth2004logic}. 
Interactive Theorem Provers (ITPs) like Coq~\citep{coq1996coq}, Isabelle/HOL~\citep{nipkow2002isabelle}, and Lean~\citep{de2015lean,moura2021lean} are cornerstone tools in this domain, enabling mathematicians and engineers to construct machine-checkable proofs. 
However, the process of formalization and proof construction often requires significant expertise and manual effort, creating a bottleneck for wider adoption.

The advent of reasoning capabilities in Large Language Models (LLMs) has revolutionized many fields, particularly in the field of informal mathematical reasoning~\citep{guo2025deepseek,claude2025sonnet,qwq32b}. 
Inspired by this, our prior work~\citep{zhang2025leanabell} leverages reinforcement learning (RL) with verified rewards (\textit{e.g.}, Lean 4 verifier feedback) to foster reasoning processes exhibiting variable cognitive behaviors, thus automating or assisting in formal proof generation.
Kimina-Prover-Preview~\citep{wang2025kimina} consolidates this procedure through a distinct reasoning pattern composed of consecutive blocks integrating informal reasoning steps and Lean 4 code snippets.  
DeepSeek-Prover-V2~\citep{ren2025deepseek} further develops a thinking pattern suitable for complex theorems through subgoal decomposition.

However, a critical challenge persists: the sample efficiency in formal reasoning still lags behind that of informal reasoning models. Formal reasoning models are often evaluated by a large sampling budget (pass@k, k=32→1024) demands prohibitively large sampling budgets. Meanwhile,  practical RL training rarely exceeds 8–32 rollouts due to computational constraints and training efficiency.  Consequently, effective RL training requires an exceptionally capable LLM backbone that can operate within constrained sampling budgets and maintain reward diversity in the rollouts.
This necessity is reflected in Kimina-Prover-Preview and DeepSeek-Prover-V2, which employ the Qwen-2.5-72B~\citep{qwen2.5} and DeepSeek-Prover-V2-671B~\citep{ren2025deepseek} models respectively for RL training. However, smaller and weaker LLMs (such as those in 7B sizes) generally lack the capacity to satisfy this requirement under standard RL training configurations.

In this work, we aim to enhance the formal reasoning performance of small-size LLMs. We observe that these small-size models often fail to generate high-quality reasoning steps or corresponding Lean 4 codes with sufficient reliability in a small sampling budget. To address this limitation, we integrate the Lean 4 code execution as a service use through an instant, multi-turn interaction paradigm. Our reasoning process may involves consecutive verification feedback from the formal system itself to iteratively refine outputs.
Specifically, whenever the model generates a Lean 4 snippet—whether within an intermediate reasoning chain or as part of the final solution—we immediately execute and verify it using the Lean 4 verifier. The resulting feedback, which ranges from successful proof-state advancement to type errors or tactic failures, is translated into reward signals that either guide subsequent reasoning steps or directly compute policy gradients.
We train our model with this verifier integrated reasoning pattern end-to-end via RL (\textit{e.g.}, 
DAPO~\citep{yu2025dapo}), effectively guiding the policy to prefer actions that lead to verifiable proof steps and to avoid those that result in errors. 

Our experiments show that verifier-integrated long chain-of-thought reasoning significantly enhances the model’s capabilities in formal theorem proving. The resulting Leanabell-Prover-V2 model with 7B parameters establishes a new state-of-the-art in neural theorem proving with a small-size LLM across multiple benchmarks. On MiniF2F-test, it achieves 78.2\% accuracy on pass@128, surpassing the DeepSeek-Prover-V2-7B (CoT) by a margin of 2\%. The model demonstrates strong generalization capabilities to college-level theorem proving, achieving a 6.4\% improvement over the Kimina-Prover-Preview-Distill-7B baseline at pass@128 on ProofNet-test problems. While Leanabell-Prover-V2 does not exceed DeepSeek-Prover-V2-7B's overall performance on ProverBench, it successfully solves one additional AIME 24\&25 problem, indicating the potential for complex mathematical reasoning tasks.

\section{Method}
\label{sec:method}

\subsection{Problem Formulation}
\label{subsec:formulation}

A recent trend involves empowering LLMs to use external tools to overcome their intrinsic limitations (\textit{e.g.}, performing calculations, accessing real-time information, executing code)~\citep{parisi2022talm,schick2023toolformer,yao2023react,shinn2023reflexion,patil2024gorilla,dong2024self,ziegenbein2024llm,jin2025search,song2025r1,wang2025otc,feng2025retool,qian2025toolrl}. 
These approaches often use RL or sophisticated prompting strategies to learn tool-using policies. Our work falls squarely within this paradigm. We conceptualize the Lean 4 verifier as a specialized tool that the LLM learns to interact with. The ``strategy'' the LLM learns via RL is how to generate proof steps that are acceptable to this verification tool. Unlike general-purpose tool use, our tool (\textit{i.e.}, Lean 4 verifier) provides highly structured, state-dependent feedback directly related to the core reasoning task, allowing for a tight feedback loop focused on logical validity and enabling the model to learn self-correction capabilities for formal proofs.

We first formulate the problem of verifier-integrated theorem proving. Given an LLM-based theorem proving model $\mathcal{M}_\theta$ and a formal theorem statement $s \in \mathcal{S}$, where $\theta$ refers to the model parameters, we can sample a set of generated proofs $\mathcal{P}= \mathcal{M}_\theta(s) =\{p_1, p_2, \cdots, p_N\}$. After feeding these generated proofs to the Lean 4 verifier, we can obtain the corresponding correctness labels $\mathcal{Y}=\{y_1, y_2, \cdots, y_N\}$ where $y_i\in\{0, 1\}$. Then, we denote $\mathcal{O}=\{o_1, o_2, \cdots, o_N\}$ as the corresponding feedbacks from the Lean4 verifier for the generated proofs.
We optimize $\mathcal{M}_\theta$ by maximize the conditional probability:
\begin{equation}
    \max_{\theta}\mathbb{E}_{s\sim\mathcal{S}, {p_i}\sim\mathcal{P}}P(y_i = 1)
    \label{eq:standard_prover}
\end{equation}
where $y_i= \mathbb{1}(p_i|s)$, and $\mathbb{1}(\cdot)$ refers to an indicator function that represents whether the generated proof $p_i$ can pass verification by the Lean 4 verifier, given the formal theorem statement $s$. %

When we introduce verifier-integrated procedure during the long CoT process and provide validation feedback to the model, the optimization process becomes:
\begin{equation}
    \max_{\theta}\mathbb{E}_{s\sim\mathcal{S}, {p_i}\sim\mathcal{P}, o_i\sim\mathcal{O}}P(\hat{y}_i = 1)
    \label{eq:tool_prover}
\end{equation}
where $\hat{y}_i = \mathbb{1}(\hat{p}_i |p_i, o_i, s)$, and $\hat{p}_i$ is the rewritten version of $p_i$ according to the validation feedback $o_i$. Therefore, compared to the optimization objective Eq.~(\ref{eq:standard_prover}), the focus here is on enabling the model to learn how to correct erroneously generated proofs after invoking Lean 4 verifier to obtain feedback, which better demonstrates the model's reflection and error-correction capabilities.

\subsection{Cold-start for Verifier-integrated Theorem Proving Generation}
\label{subsec:cold_start}

\noindent\textbf{Cold-start Data.} In order to equip the model with the ability to generate formatted Chain-of-thought (CoT) outputs for verifeir calling, we construct specially styled long CoT data for cold-start training. In the long CoT data, we employ special markers (\textit{e.g.}, \texttt{<code></code>}, \texttt{<interpreter></interpreter>}) to explicitly delineate the generated Lean 4 code blocks and verifier invocation triggers, and verifier feedback information. Figure~\ref{fig:tool-integrated-prompt} shows an example prompt for verifier-integrated theorem proving.

\begin{figure}[ht]
    \centering
    \begin{center}
    \begin{promptbox}[Prompt for Verifier-integrated Theorem Proving]
    Think step by step and solve the following problem in Lean 4. When you think you've come up with the final answer, put your Lean 4 code in <code> and </code>. The code will be compiled in the Lean 4 compiler. The compilation results will be placed between <interpreter> and </interpreter>. Then, revise your answer based on the compilation log and provide the final Lean 4 code.\newline
    
    \# Problem: Calculate the value of $11^5 - 5 \cdot 11^4 + 10 \cdot 11^3 - 10 \cdot 11^2 + 5 \cdot 11 - 1$. The answer is 100000.\newline 
    \# Formal statement:\newline
    \begin{lstlisting}[frame=single]
```lean4
import Mathlib 
import Aesop 
set_option maxHeartbeats 0 
open BigOperators Real Nat Topology Rat
    
/-- Calculate the value of $11^5 - 5 \cdot 11^4 + 10 \cdot 11^3 - 10 \cdot 11^2 + 5 \cdot 11 - 1$. Show that it is 100000.-/
    
theorem my_favorite_theorem : 11^5 - 5 * 11^4 + 10 * 11^3 - 10 * 11^2 + 5 * 11 - 1 = 100000 := by
'''\end{lstlisting}
    \end{promptbox}
    \end{center}
    \caption{Prompt for verifier-integrated theorem proving, where we encourage the model employ Lean 4 verifier to verify the correctness of the generated proofs.}
    \label{fig:tool-integrated-prompt}
\end{figure}

There are four main steps during the automatic construction of the cold-start data:
\begin{itemize}
    \item \textit{Straightforward Proof Generation.} For an input theorem statement, we leverage the existing prover LLMs (\textit{e.g.}, \kimina~\citep{wang2025kimina} and \dsii~\citep{ren2025deepseek}) to collect multiple generated proofs.  
    \item  \textit{Proof Validation.} We employ the Lean 4 verifier to strictly verify the generated proofs and filter out incorrect proofs that fail validation.
    \item \textit{Automatic Correction.} Given the incorrect proofs and their corresponding errors provided by the Lean 4 verifier, we design a special prompt (See Figure~\ref{fig:correcting-prompt} in Appendix~\ref{app:cold_start}) to ask the Claude-3.7-Sonnet~\citep{claude2025sonnet} to rewrite the proofs. 
    \item \textit{Re-verification.} We perform the Lean verification again on the rewritten results. Once we obtain rewritten proofs that pass verification, we successfully construct a set of high-quality training data in the form of ``incorrect-corrected'' pairs.
\end{itemize}

\begin{figure}[ht]
    \centering
    \begin{center}
    \includegraphics[width=0.85\textwidth]{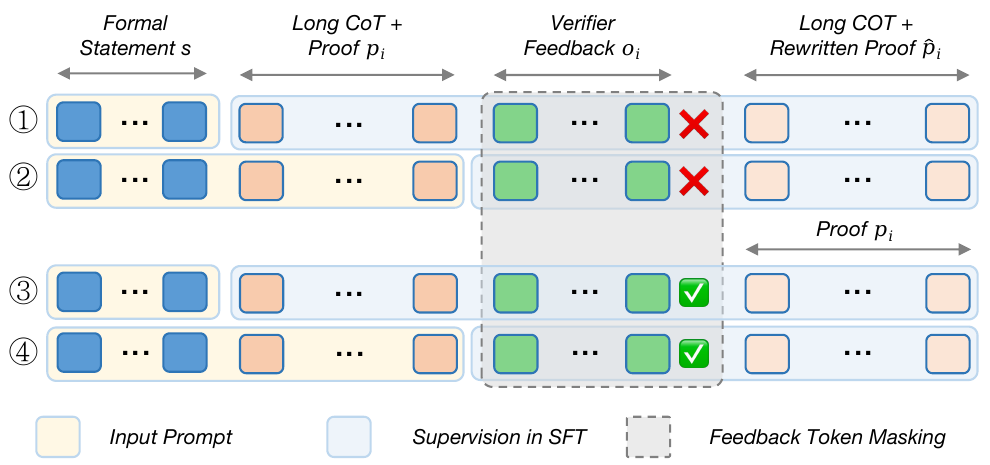}
    \caption{Overview of four scenarios to cold start data synthesis.}
    \label{fig:cold-start-data}
    \end{center}
\end{figure}

After collecting the ``incorrect-corrected'' pairs, we consider four scenarios to synthesize data with long CoTs, as shown in Figure~\ref{fig:cold-start-data}, aiming to achieve two goals: on one hand, enabling the model to explicitly output the thinking process before generating proofs, and on the other hand, when verifying generated proof failed, combining verification feedback information to learn to reflect and correct the generated results. Here are the four scenarios:
\begin{itemize}
    \item For the case ``$p_i$ failed but $\hat{p}_i$ success'', we take the formal statement $s$ as input, concatenate the long COT and original proof $p_i$ with the rewritten $\hat{p}_i$ as the output. In this scenario, we collect around 2K samples.
    \item For the case ``$p_i$ failed but $\hat{p}_i$ success'', we concatenate the formal statement $s$ with the long CoT and the original proof $p_i$ as input, regard  the rewritten $\hat{p}_i$ as the output. In this scenario, we obtain around 2K samples.
    \item For the case ``$p_i$ success'', we use the formal statement $s$ as input, and regard the long CoT and the original proof $p_i$ as output. In this scenario, we collect around 2K samples.
    \item For the case ``$p_i$ success'',  we use the formal statement $s$ and the long CoT as input, and the original proof $p_i$ as output. In this scenario, we collect around 1K samples. 
\end{itemize}
We refer the reader to Figure~\ref{fig:cold-start-prompt} in Appendix for additional details on the used prompt for the first two scenarios. Since in the latter two scenarios, we can directly obtain results with long CoT through concatenation methods, there is no need for special prompts for generation.
Finally, we collect cold-start data for Kimina-Prover-Preview-Distill-7B~\citep{wang2025kimina} and  DeepSeek-Prover-V2-7B~\citep{ren2025deepseek} respectively. All of these samples are either ``incorrect-to-correct'' data or ``correct'' data with synthesized long CoTs. 

\noindent\textbf{Feedback Token Masking.}  Given that Lean 4 verification information is not generated by the model itself, we apply a feedback token masking strategy that focuses solely on predicting the rewritten portions, thereby reducing potential interference during the supervised fine-tuning (SFT) stage. As shown in Figure~\ref{fig:cold-start-data}, we mask the token sequences provided by the Lean 4 verifier in the four scenarios of the synthesized cold-start data. 

\noindent\textbf{Training Setup.} We employ two existing powerful models as our baselines, including Kimina-Prover-Preview-Distill-7B~\citep{wang2025kimina} and DeepSeek-Prover-V2-7B~\citep{ren2025deepseek}. These two models are not able to utilize Lean 4 verifier to correct the generated proofs within the long CoTs. 
During the cold-start SFT training, we set the batch size to 128, the learning rate to 2e-5, and trained for three epochs. After this lightweight training, we examine the format of the model's output and find that the model already satisfy the corresponding formatting requirements. This indicates that the cold start training process is sufficient.

\begin{figure}[ht]
    \centering
    \begin{center}
    \includegraphics[width=\textwidth]{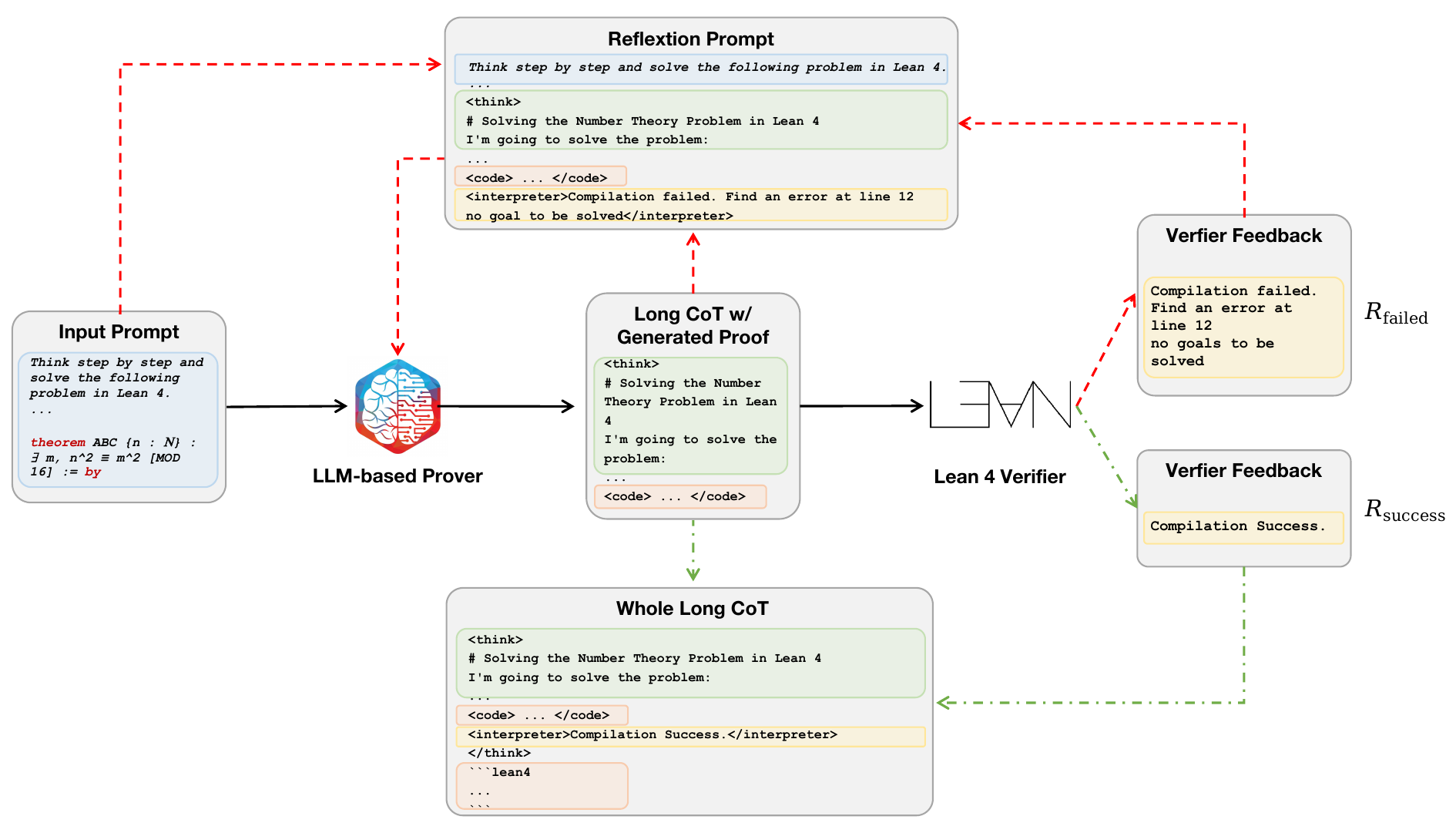}
    \caption{Overview of our proposed verifier-integrated theorem proving framework.}
    \label{fig:rl-framework}
    \end{center}
\end{figure}

\subsection{Verifier-integrated Theorem Proving via Reinforcement Learning}
\label{subsec:too_use}

We integrate Reinforcement Learning (RL) with the Lean 4 theorem prover and Lean 4 verifier to automate the generation and correction of formal proofs. 
The RL agent interacts with the Lean 4 environment, generating whole proofs and receiving feedback from Lean 4 verifier as reward signals. %

\noindent\textbf{RL Framework.} As shown in Figure~\ref{fig:rl-framework}, we first provide the formal statement $s$ that needs to be solved and the format requirements for the solving process as input prompt to the LLM-based prover. The prover model, after cold-start training, can generate long CoT reasoning according to the requirements, and provide an initial proof within the long CoT, placing it between \texttt{<code></code>} delimiters. When the generated initial proof is input to the Lean 4 verifier, there are two possible feedback results: compilation success or compilation failure. 

For cases where compilation is successful, we can return the proof from the already generated long CoT as the correct result. However, for compilation failed cases, we need to place the compilation error information between \texttt{<interpreter></interpreter>} delimiters, then concatenate it with the previous round's long COT to serve as a reflection prompt for the prover model, acting as input for a new round regeneration. Additionally, we will calculate and provide rewards $R_{\text{success}}$ and $R_{\text{failed}}$ respectively in the RL training process, based on whether the compilation is correct or incorrect. 

\noindent \textbf{Policy Optimization Algorithm.} We employ the recent DAPO~\citep{yu2025dapo} as our RL algorithm, which has been proven to be highly effective on solving several important issues such as entropy collapse, reward noise, and training instability. DAPO samples a group of output proofs $\{p_i\}_{i=1}^G$ for each formal statement $s$, and optimizes the policy for formal proof generation via the following objective:
\begin{equation}
\begin{aligned}
\mathcal{J}_{\text{DAPO}}(\theta) =\quad& \mathbb{E}_{s\sim \mathcal{S}, \{p_i\}_{i=1}^G\sim \pi_{\theta_\text{old}}(\cdot\mid q)}\\&
\Bigg[\frac{1}{\sum_{i=1}^{G}|p_i|}\sum_{i=1}^{G}\sum_{t=1}^{|p_i|} 
\min \Big( r_{i,t}(\theta) \hat{A}_{i,t},  
\ \text{clip} \Big( r_{i,t}(\theta), 1 - {\varepsilon_{\text{low}}}, 1 + {\varepsilon_{\text{high}}} \Big) \hat{A}_{i,t} \Big) \Bigg]
\\
\text{s.t.}\quad& 0< \Big|\{p_i\mid\texttt{is\_valid}(p_i)\}\Big|< G,
\label{eq:dapoloss}
\end{aligned}
\end{equation}
where \texttt{is\_valid()} refers to the validation status provided by the Lean4 verifier, $\varepsilon_\text{low}$ and $\varepsilon_\text{high}$ denote the lower and upper clipping ratios, empirically set to 0.2 and 0.28, respectively. $r_{i,t}$ and $\hat{A}_{i,t}$ are formulated respectively as follows:
\begin{equation}
    r_{i,t}(\theta)=\frac{\pi_{\theta}(p_{i,t} \mid q, p_{i,<t})}{\pi_{\theta_{\text{old}}}(p_{i,t} \mid q,p_{i,<t})},\quad\hat{A}_{i,t} = \frac{R_i - \text{mean}(\{R_i\}_{i=1}^G)}{\text{std}(\{R_i\}_{i=1}^G)}.
\label{eq:advantage_calculation}
\end{equation}
Notably, $R_i$ refers to the reward for the output proof $p_i$. During our training, we remain the strategies used in the original DAPO algorithm, inlcuding clip-higher, 
and token-level policy gradient loss. %
Similar to the cold-start fine-tuning, we also mask the token sequences provied by the Lean 4 verifier for stable RL training.

\noindent\textbf{Reward Design.} In our experiments, we primarily use two rewards: format reward (\textit{i.e.}, $R_\text{format}$) and compilation status reward (\textit{i.e.}, $R_\text{failed}$ and $R_\text{success}$). The $R_\text{format}$ is set to a smaller value (such as 0.2) compared to $R_\text{failed}$/$R_\text{success}$ (such as 1.0).  For assessing the generated proof quality, we find that employing the simple $R_\text{failed}$ and $R_\text{success}$ are sufficient for the efficient RL training. Based on the detailed verification feedback obtained, we can construct a structured reward signal. As shown in the Appendix~\ref{app:rewards-design}, we have investigated such method, unfortunately, we have not obtained favorable results. This leaves behind a crucial question: whether we can design elaborate rewards based on Abstract Syntax Tree (AST) feedback (See Figure~\ref{fig:ast-feedback} in Appendix) to help models learn effective reflection capabilities in RL training.

\noindent\textbf{Training Setup.} We use NuminaMath~\citep{numina_math_datasets} as source to collect the RL training data. We translate the natural language math problems into formal statements by utilizing the public Kimina-Autoformalizer-7B~\citep{wang2025kimina}. For each formal statement, we use the supervised finetuned model, initialized with the cold-start data, to generate multiple proof candidates and compute the pass rate (\textit{i.e.}, pass@8). Subsequently, we select formal statements achieving pass rates in the range of 1/8 to 1/2 to constitute our training dataset. For the Kimina-Prover-Distill-7B~\citep{wang2025kimina} model, this filtering process results in a curated training dataset of 3.8K valid formal statements. Training hyperparameters are configured as follows: maximum token length = 16K, batch size = 128, rollout size = 24, learning rate = 1e-6, and temperature = 0.6. Similarly, for the DeepSeek-Prover-V2-7B~\citep{ren2025deepseek}, this filtering process results in a curated training dataset of 1.7K valid formal statements. Training hyperparameters are configured as follows: maximum token length = 16K, batch size = 128, rollout size = 24, learning rate = 1e-6, and temperature = 0.7.

\section{Experimental Results}
\label{sec:experiments}
In this section, we present the main experimental results after the SFT and RL stages with the two baseline models, including Kimina-Prover-Preview-Distill-7B~\citep{wang2025kimina} and DeepSeek-Prover-V2-7B~\citep{ren2025deepseek}. All experimental results are conducted with Lean 4.9.0, using the same testing environment as the DeepSeek Prover series~\citep{xin2024deepseek-v1.5,ren2025deepseek}. 

\begin{figure}[t]
\begin{minipage}{0.325\textwidth}
    \includegraphics[width=\textwidth]{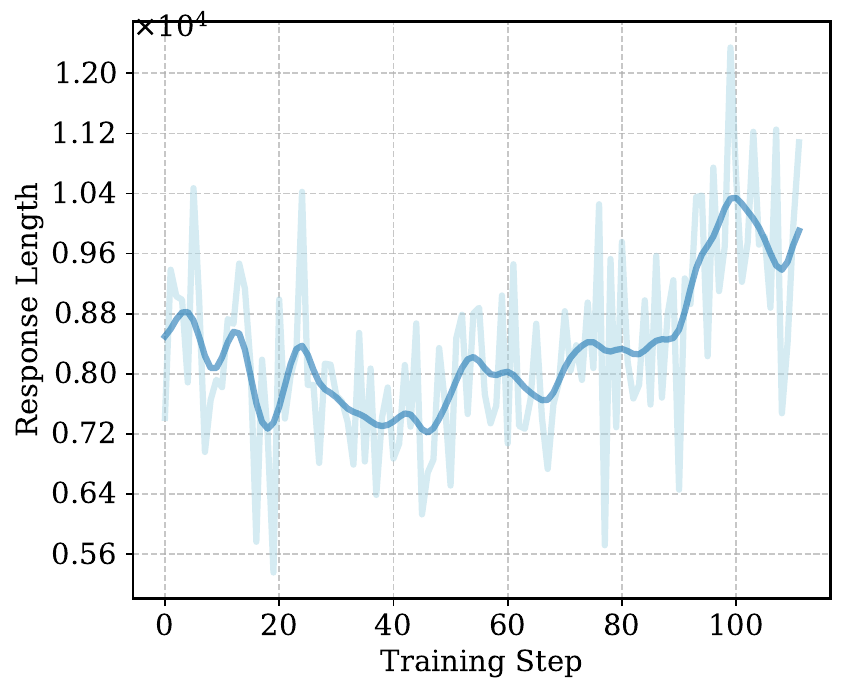}
\end{minipage}
 \hfill
\begin{minipage}{0.325\textwidth}
    \centering
    \includegraphics[width=\textwidth]{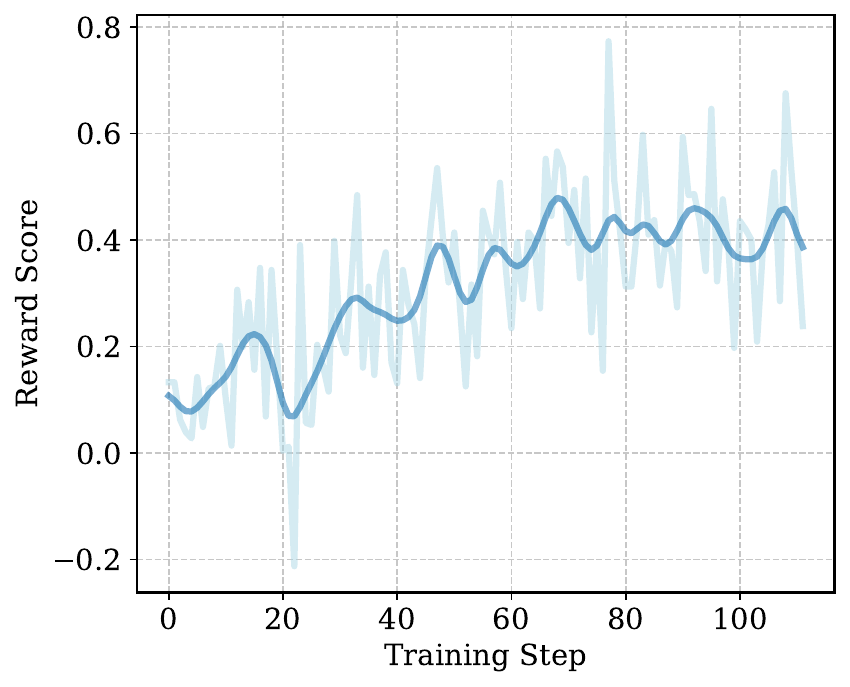}
\end{minipage}
\hfill
\begin{minipage}{0.325\textwidth}
    \centering
    \includegraphics[width=\textwidth]{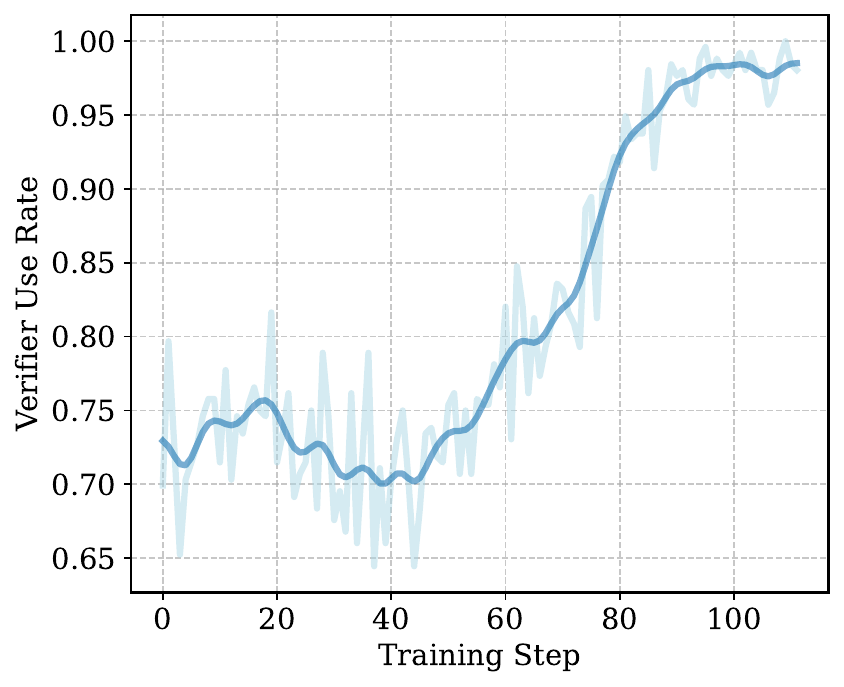}
\end{minipage}
\hfill
\begin{minipage}{0.325\textwidth}
    \centering
    \includegraphics[width=\textwidth]{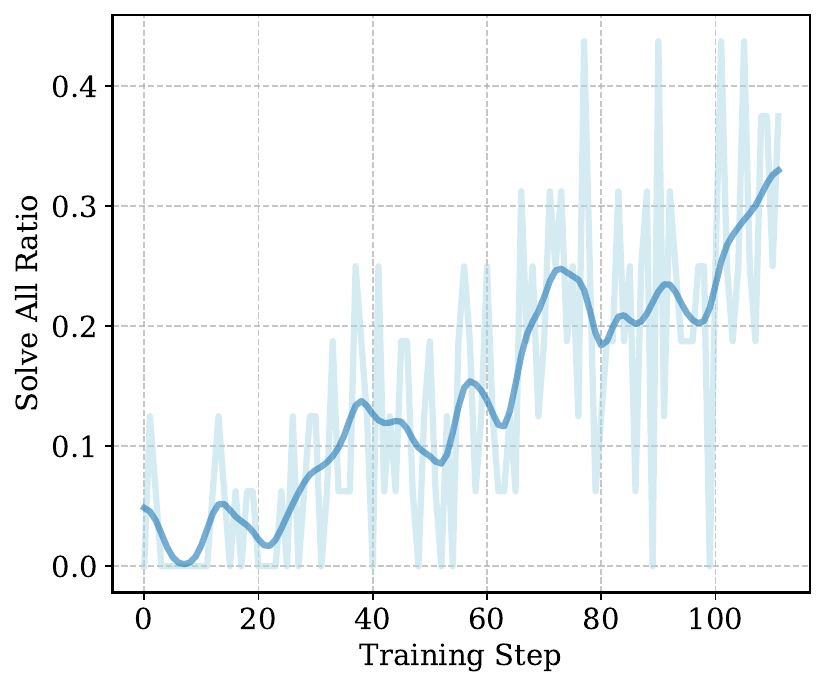}
\end{minipage}
\hfill
\begin{minipage}{0.325\textwidth}
    \centering
    \includegraphics[width=\textwidth]{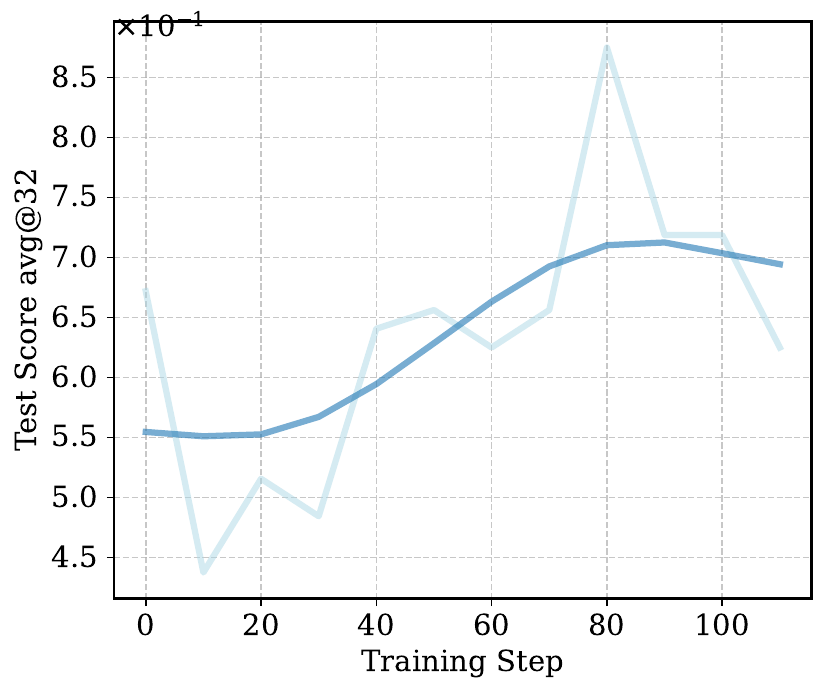}
\end{minipage}
\hfill
\begin{minipage}{0.325\textwidth}
    \centering
    \includegraphics[width=\textwidth]{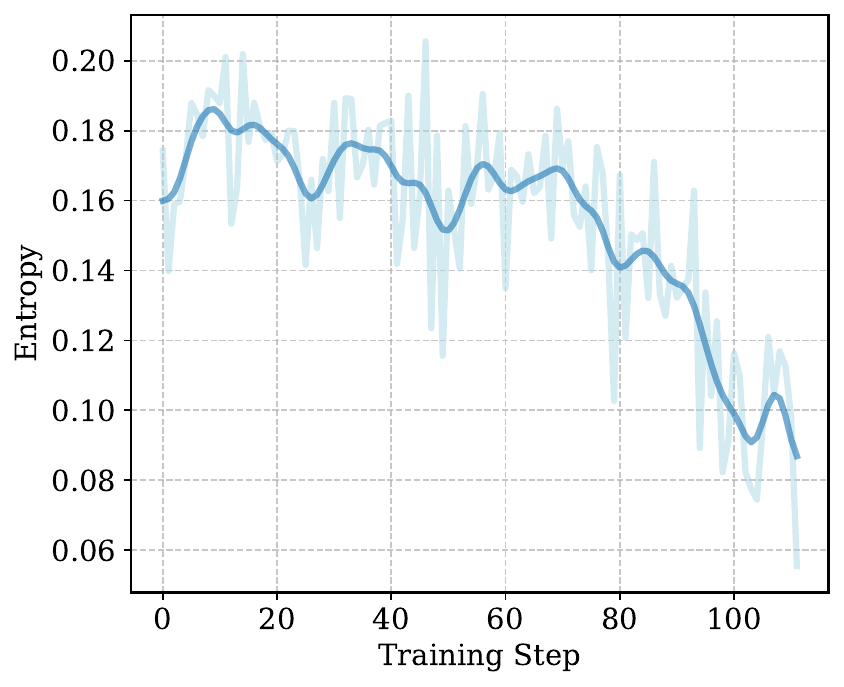}
\end{minipage}
\caption{Several core records in the RL training with  DeepSeek-Prover-V2-7B model, including \textit{Response Length}, \textit{Reward Score}, \textit{Verifier Use Rate}, \textit{Solve All Ratio}, \textit{Test Score avg@32}, and \textit{Entropy}.}
\label{fig:rl-records}
\end{figure}

In Figure~\ref{fig:rl-records}, we present the RL training records on the DeepSeek-Prover-V2-7B model. We can see that the response length shows a slight increase as training steps progress, which is related to the token-level loss introduced by DAPO~\citep{yu2025dapo}, and this aligns with the official experimental observations. %
Looking at the two charts showing reward score and solve all ratio together, we find that the model continuously receives more positive rewards during the training, with an increasing number of ``solve all'' problems appearing in the training set.
This indicates that the RL training scheme has improved the model's problem-solving capabilities. The averaged test score at pass@32 with our validation set, which is a small subset sampled from our training data, also shows an upward trend, reflecting that the training is effective.
In addtion, since we encourage the model to use verifier validation in the training prompts, the ``verifier use rate'' shows an upward trend during the training process, and after brief training, almost all problems utilize verifier validation. Finally, during the training process, entropy shows a slow declining trend without the occurrence of entropy collapse. For a comparison, we refer the authors to the Appendix~\ref{app:rl-algoritm} for more details, where we provide the training records with a GRPO~\citep{shao2024deepseekmath} algorithm that uses a different token-level loss calculation strategy. 

As shown in Figure~\ref{fig:success_example-1} and Figure~\ref{fig:success_example-2} in the Appendix~\ref{app:success_example}, we present an example that our proposed verifier-integrated long CoT reasoning helps the model to correct the errors in the generated proof. In the following part of this section, we present detailed quantitative comparison results over multiple benchmarks.  

\subsection{Results on MiniF2F Benchmark}
\label{subsec:results-on-minif2f}
\noindent\textbf{Main Results.} We first validate the effectiveness of our proposed method on the most commonly-used MiniF2F-test~\citep{zheng2021minif2f} benchmark. We have two versions of our models post-trained from two strong prover models: Kimina-Prover-Preview-Distill-7B and Deepseek-Prover-V2-7B, namely Leanabell-Prover-V2-KM and Leanabell-Prover-V2-DS.
We mainly compare current whole proof generation methods, while ignore those with proof-step methods using far more inference-compute.
As shown in Table~\ref{tab:minif2f_results}, our posttraining framework boosts both Kimina-Prover-Preview-Distill-7B and DeepSeek-Prover-V2-7B models. First, our method boosts the baseline Kimina-Prover-Preview-Distill-7B by a large margin of 5.3\% (pass@32). Remarkably, we reach equivalent performance levels, where our method with pass@128 achieves the comparable performance as the baseline with pass@1024 (\textit{i.e.}, 70.4\% v.s. 70.8\%), demonstrating considerable inference efficiency gains.  
Meanwhile, we obatain a 1\% improvement over the baseline DeepSeek-Prover-V2-7B at pass@32, with the advantage growing larger as the sample budget increases, ultimately achieving a 2\% improvement over the baseline at pass@128.

\noindent\textbf{RL Ablation.} In Table~\ref{tab:rl-ablation}, we discover that our proposed verifier-integrated reasoning method yields benefits on both base models Kimina-Prover-Preview-Distill-7B and DeepSeek-Prover-V2-7B. Furthermore, the improvements are more substantial compared to employing the vanilla RL method that explores on all plain text CoTs. First, we observe that applying the vanilla RL method on Kimina-Prover-Preview-Distill-7B yields 2.4\% improvement, while our verifier-integrated RL method achieves a significant 5.3\% improvement at pass@32. Second, we discover that the vanilla RL method achieves no gain on DeepSeek-Prover-V2-7B, while our method can survive to obtain 1\% improvement at pass@32.

\begin{figure}[ht]
    \centering
    \begin{center}
    \includegraphics[width=0.7\textwidth]{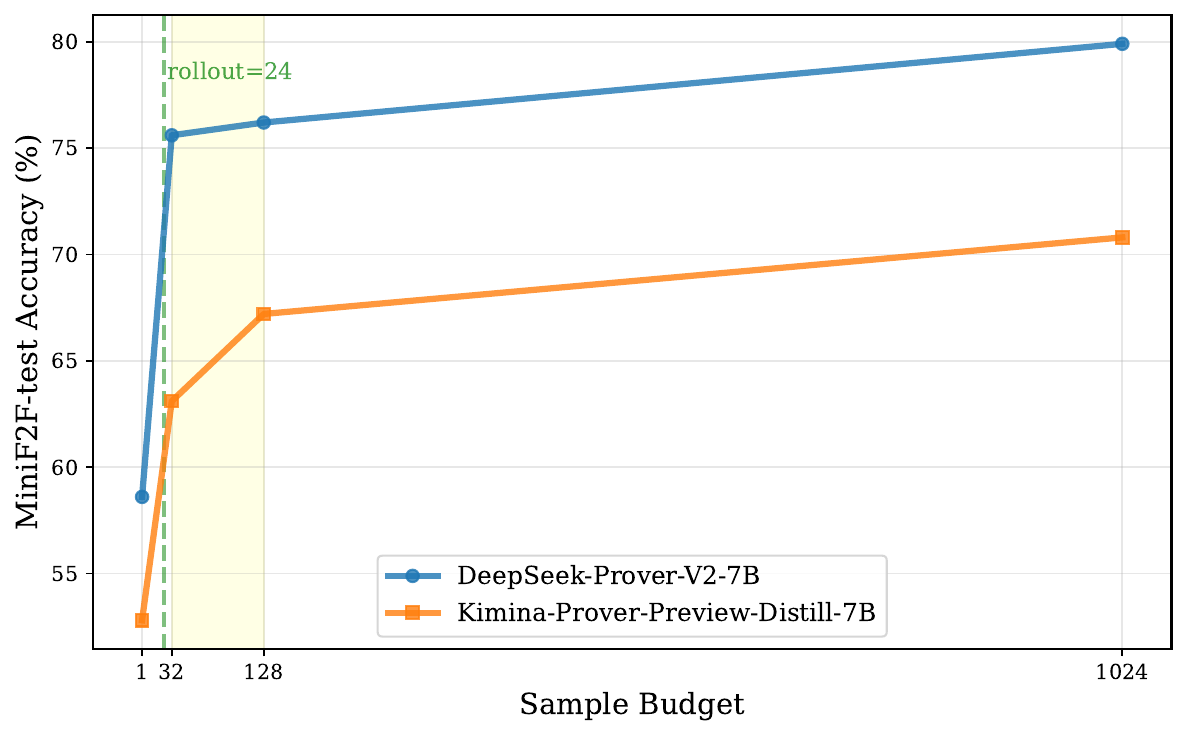}
    \caption{The curves of Kimina-Prover-Preview-Distill-7B and DeepSeek-Prover-V2-7B models' accuracy on MiniF2F-test as the sample budget varies.}
    \label{fig:pass_at_k_slope}
    \end{center}
\end{figure}

Figure~\ref{fig:pass_at_k_slope} shows the curves of Kimina-Prover-Preview-Distill-7B and DeepSeek-Prover-V2-7B models' accuracy on MiniF2F-test as the sample budget varies. We can observe that since the RL training phase uses a rollout size of 24, which is relatively close to the sample budget range of 32-128. In this range, the pass@k curve of Kimina-Prover-Preview-Distill-7B has a steep slope, which is more conducive to RL stimulating exploration capabilities. In contrast, the curve of DeepSeek-Prover-V2-7B has a flat slope in this range, the exploration capability it can stimulate is limited.
We analyze that this is because DeepSeek-Prover-V2-7B is obtained through comprehensive large model distillation and subsequent RL enhancement. The previous RL training has already substantially activated the model's problem-solving abilities, making it very challenging to achieve significant gains through an additional RL training. %

\noindent\textbf{On the Verifier Calling Iteration.}  Although we train the model to perform verifier feedback-based reflection only once per long CoT during training, in practice, we can verify after each generated \texttt{<code></code>} block in the long CoT and have the model reflect iteratively until it produces correct proofs or hits the maximum iteration threshold. 
In Table~\ref{tab:verifier-iterations}, we demonstrate the effectiveness of verifier calling iteration cycles, finding that the gains are quite significant during the first round of reflective modifications based on the validation feedback. For example, Leanabell-Prover-V2-KM improves from 64.7\% to 68.4\% (\textit{i.e.}, an increase of 3.7\%), and Leanabell-Prover-V2-DS improves from 75.4\% to 76.6\% (\textit{i.e.}, an increase of 1.2\%). When continuously increasing the iteration cycles from 1 to 3, we can also achieve an improvement by a margin of 0.8\%. The results indicate promissing potential, suggesting Leanabell-Prover-V2-KM may achieve additional gains with more increased iterations. Unfortunately, we do not observe improvements on Leanabell-Prover-V2-DS. 

In addition, we have explored to increase the rollout size in each iteration cycle (\textit{i.e.}, expanding into a tree-like branching structure). For example, when the rollout size is increased from 1 to 4 within an iteration cycle, we can achieve a slight improvement of 0.4\% (\textit{i.e.}, 68.4\% $\to$ 68.8\%) with Leanabell-Prover-V2-KM. Similarly, we do not obtain improvements with Leanabell-Prover-V2-DS. Limited by computational resource constraints, we do not evaluate larger ``\#'' values. Nevertheless, results from Leanabell-Prover-V2-KM indicate promising potential for additional performance gains via increased tree branching width.

\begin{table*}[htp]
\setlength{\tabcolsep}{0.2in}
\begin{center}
\small
\begin{tabular}{lcc}
\toprule
    \textbf{Method} & \textbf{Sample budget} & \textbf{MiniF2F-test} \\
    \toprule
    TheoremLlama [\citenum{wang2024theoremllama}] & 128 & $33.6\%$ \\
    \midrule
    DeepSeek-Prover-v1 [\citenum{xin2024deepseek-v1}] & $128$ & $46.1\%\pm 0.5\%$ \\
    \midrule
    \multirow{3}{*}{\dsi-Base~[\citenum{xin2024deepseek-v1.5}]}  & 128 & $29.7\%\pm 0.5\%$ \\
    & 3200 & $39.2\%$ \\
    & 6400 & $42.2\%$ \\
    \midrule
    \multirow{4}{*}{DeepSeek-Prover-v1.5-SFT~[\citenum{xin2024deepseek-v1.5}]}  & 32 & $48.2\% \pm 0.6\%$ \\
     & 64 & $49.6\% \pm 0.7\%$ \\
     & $128$ & $50.4\% \pm 0.4\%$ \\
     & $3200$ & $53.3\% \pm 0.5\%$ \\
    \midrule
    \multirow{4}{*}{DeepSeek-Prover-v1.5-RL~[\citenum{xin2024deepseek-v1.5}]}  & 32 & $50.0\%\pm 0.5\%$ \\
     & 64 & $50.7\%\pm 0.4\%$ \\
     & $128$ & $51.6\%\pm 0.5\%$ \\
     & $3200$ & $54.9\%\pm 0.7\%$ \\
    \midrule
    \multirow{2}{*}{STP~[\citenum{dong2025stp}]}   & $128$ & $57.7\%\pm 0.6\%$ \\
     & $3200$ & $61.7\%\pm 0.6\%$ \\
    \midrule
    \multirow{2}{*}{\goedelii~[\citenum{lin2025goedel}]}  & 32 & $57.6\%\pm 0.7\%$ \\
     & $3200$ & 62.7\% \\
     \midrule
    \multirow{3}{*}{\kylean-GD-RL~[\citenum{zhang2025leanabell}]} & 32 & 59.8\% \\
     & 64 & 60.7\% \\
     & $128$ & 61.1\% \\
     \midrule
    \multirow{3}{*}{Kimina-Prover-Preview-Distill-7B~[\citenum{wang2025kimina}]}
     & 32 & 63.1\% \\
     & 128 & 67.2\% \\
     & 1024 & 70.8\% \\
     \midrule
    \multirow{3}{*}{DeepSeek-Prover-V2-7B (CoT)~[\citenum{ren2025deepseek}]} &  32 & 75.6\% $\pm$ 0.5\% \\
    & 128 & 76.2\% \\
     & 1024 & 79.9\% $\pm$ 0.3\% \\
     \midrule
    \multirow{2}{*}{\kylean-V2-KM (Ours)} &  32 & 68.4\% \\
     & 128 & 70.4\% \\
     \midrule
    \multirow{2}{*}{\kylean-V2-DS (Ours)} &  32 & 76.6\% \\
     & 128 & 78.2\% \\
    \bottomrule
\end{tabular}
\caption{
Comparison with state-of-the-art methods on the miniF2F-test dataset. The notation $\mu\pm\sigma$ denotes the average accuracy $\mu$ and the standard deviation $\sigma$. ``KM'' and ``DS'' refer to using the Kimina-Prover-Preview-Distill-7B and DeepSeek-Prover-V2-7B as base models. Our proposed method employs a rollout size of 1 for subsequent Lean 4 verifier calling iteration cycles.
}
\label{tab:minif2f_results} %
\end{center}
\end{table*}

\begin{table*}[htp]
\setlength{\tabcolsep}{0.2in}
\begin{center}
\small
\begin{tabular}{lccccc}
\toprule
    \textbf{Method} & \textbf{Sample budgets} & \textbf{MiniF2F-test} \\
    \midrule
    Kimina-Prover-Preview-Distill-7B~\citep{wang2025kimina}  & \multirow{3}{*}{32} & 63.1\% \\
     \quad + Vanilla RL& & 65.5\% \\
     \quad + Our RL & & 68.4\%\\
    \midrule
    DeepSeek-Prover-V2-7B~\citep{ren2025deepseek} & \multirow{3}{*}{32} & 75.6\% \\
     \quad + Vanilla RL& & 75.6\% \\
     \quad + Our RL & & 76.6\% \\
    \bottomrule
\end{tabular}
\caption{
Comparison between direct RL method (\textit{i.e.}, without using validation feedback information) and our proposed verifier-integrated reasoning method. For a fair comparison, both policy optimization algorithm and reward are with the same configurations.
}
\label{tab:rl-ablation} %
\end{center}
\end{table*}

\begin{table*}[htp]
\setlength{\tabcolsep}{0.2in}
\begin{center}
\small
\begin{tabular}{lccccc}
\toprule
    \multirow{2}{*}{\textbf{Method}} & \multirow{2}{*}{\textbf{Sample budgets}} & \multicolumn{3}{c}{\textbf{Iterations}} \\ \cmidrule(lr){3-5}
    & & 1 & 2 & 3  \\
    \midrule
    \multirow{3}{*}{\kylean-V2-KM (Ours)} & 32 (Vanilla) & 64.7\% & - & - \\
    & 32-1 & 68.4\% & 68.8\% & 69.2\% \\
     & 32-4 & 68.8\% & - & - \\
     \midrule
    \multirow{3}{*}{\kylean-V2-DS (Ours)} & 32 (Vanilla) & 75.4\% & - & - \\
     & 32-1 & 76.6\% & 76.6\% & 76.6\% \\
     & 32-4 & 76.6\% & - & -  \\
    \bottomrule
\end{tabular}
\caption{
Evaluation across diverse budget configurations and Lean 4 verifier calling iteration cycles on MiniF2F-test. ``32-\#" refers to 32 rollouts in the first round, followed by \# rollouts in each subsequent Lean 4 verifier calling iteration. ``Vanilla'' refers to evaluating the generated proof in the first \texttt{<code></code>} block in the long CoTs.
}
\label{tab:verifier-iterations} %
\end{center}
\end{table*}

\subsection{Results on Undergraduate-level Benchmarks}
\label{subsec:results-on-undergradute}

\noindent\textbf{ProofNet}~\citep{azerbayev2023proofnet} consists of 371 problems in Lean 3, drawn from a range of
popular undergraduate pure mathematics textbooks. Similar to DeepSeek-Prover-V2~\citep{ren2025deepseek}, we use the Lean 4 translation of ProofNet made available in DeepSeek-Prover-V1.5~\citep{xin2024deepseek-v1.5}. We employ ProofNet-test for model evaluation, which contains 186 problems. As shown in Table~\ref{table:proofnet-putnam}, after applying our verifier-integrated reasoning method on Kimina-Prover-Preview-Distill-7B, we surpass the performance of the base model at pass@128 when using pass@32, reaching 13.4\%. Compared to Kimina-Prover-Preview-Distill-7B, Leanabell-Prover-V2-KM achieves an obvious improvement by a margin of 6.4\% at pass@128. However, on DeepSeek-Prover-V2-7B, we only achieved comparable results without significant performance improvement (Please see our RL discussion in Section~\ref{subsec:results-on-minif2f}).

\begin{table*}[htbp]
    \begin{center}
    \small
    \begin{tabular}{lcccc}
    \toprule
        \multirow{2}{*}{\textbf{Method}} & \textbf{Model}  & \textbf{Sample}  & \multirow{2}{*}{\textbf{ProofNet-test}}  \\
       & \textbf{size} & \textbf{budget} & \\
        \toprule
            Goedel-Prover-SFT~[\citenum{lin2025goedel}] & 7B & $32$ & $15.6\%$ &  \\
        \midrule
            \multirow{2}{*}{Kimina-Prover-Preview-Distill-7B~[\citenum{wang2025kimina}]} & \multirow{2}{*}{7B} &  $32$ & 10.3\%   \\
            & & $128$ & 11.8\%  \\
        \midrule
            \multirow{2}{*}{DeepSeek-Prover-V2-7B\quad(CoT)~[\citenum{ren2025deepseek}]} & \multirow{2}{*}{7B} &  $32$ & $23.0\%\pm 0.4\%$  \\
            & & $128$ & $25.4\%\pm 0.7\%$ \\
        \midrule
            \multirow{2}{*}{Leanabell-Prover-V2-KM (Ours)} & \multirow{2}{*}{7B} & $32$ & 13.4\% \\
            & & $128$ & 18.2\% \\
        \midrule
            \multirow{2}{*}{Leanabell-Prover-V2-DS (Ours)} & \multirow{2}{*}{7B} &  $32$ & 23.7\% \\
            &  & $128$ & 25.2\% \\
        \bottomrule
    \end{tabular}
    \caption{The experimental results on ProofNet-test. %
    }
    \label{table:proofnet-putnam}
    \end{center}
\end{table*}

\begin{table*}[htbp]
    \begin{center}
    \small
    \begin{tabular}{lcccc}
    \toprule
        \multirow{2}{*}{\textbf{Method}} & \textbf{Model} & \textbf{Sample}  & \multicolumn{2}{c}{\textbf{ProverBench}} \\ \cmidrule(lr){4-5}
        & \textbf{size} & \textbf{budget} & \textbf{All} & \textbf{AIME 24\&25} \\
        \toprule
            \multirow{2}{*}{STP~[\citenum{dong2025stp}]} & \multirow{2}{*}{7B} & $32$ & $27.5\%\pm 0.7\%$ & $0/15$ \\
            & & $128$ & $31.4\%\pm 1.1\%$ & $1/15$ \\
        \midrule
            \multirow{2}{*}{Kimina-Prover-Preview-Distill-7B~[\citenum{wang2025kimina}]} & \multirow{2}{*}{7B} & $32$ &  30.2\%  & 1/15  \\
            & & $128$ &  34.8\% &  1/15 \\
        \midrule
            \multirow{2}{*}{DeepSeek-Prover-V2-7B\quad(CoT)~[\citenum{ren2025deepseek}]} & \multirow{2}{*}{7B} & $32$ & $49.0\%\pm 0.3\%$ & $1/15$ \\
            & & $128$ & $50.8\%\pm 0.5\%$ & $1/15$ \\
        \midrule
            \multirow{2}{*}{Leanabell-Prover-V2-KM (Ours)} & \multirow{2}{*}{7B} & $32$ & 39.8\%  & 1/15  \\
            & & $128$ &  42.9\% &  1/15 \\
        \midrule
            \multirow{2}{*}{Leanabell-Prover-V2-DS (Ours)} & \multirow{2}{*}{7B} & $32$ & 47.8\%  & 1/15  \\
            & & $128$ &  48.7\% &  2/15 \\
        \bottomrule
    \end{tabular}
    \caption{The experimental results on ProverBench. The \texttt{All} category represents the complete evaluation set consisting of 325 problems, while AIME 24\&25 denotes a subset of 15 problems formalized from recent AIME competitions. The results for STP \citep{dong2025stp} are evaluated using the open-source model weights.
    }
    \label{table:proverbench}
    \end{center}
\end{table*}

\subsection{Results on ProverBench Benchmark}
\label{subsec:results-on-proverbench}

ProverBench~\citep{ren2025deepseek} is a benchmark dataset comprising 325 formal statements that are formalized from the recent AIME competitions (\textit{i.e.}, AIME 24 and 25), and textbook examples and educational tutorials. This benchmark is designed to
enable more comprehensive evaluation across both high-school competition problems and
undergraduate-level mathematics. The evaluation results are presented in Table~\ref{table:proverbench}. 
With Kimina-Prover-Preview-Distill-7B as the base model, our Leanabell-Prover-V2-KM achieves obvious improvements by margins of 9.6\% and 8.1\% at pass@32 and pass@128, respectively. Similar to Kimina-Prover-Preview-Distill-7B, Leanabell-Prover-V2-KM does not achieve improved performance on AIME 24\&25 problems. 
With DeepSeek-Prover-V2-7B as the base model, our Leanabell-Prover-V2-DS exhibits marginally lower performance than the base model across the entire dataset. However, we observe with surprise that it correctly solved one additional problem set among the more challenging AIME 24\&25 problems.

\section{Conclusion, Limitation, and Future Work}
\label{sec:conclusion}

In this work, we have presented a novel method to  addresses the critical limitation of small LLMs in generating reliable formal proofs by leveraging instant verification feedback within a multi-turn interaction framework.
This enables models to learn from real-time feedback, iteratively refining reasoning steps and developing preference for verifiable proof actions over error-prone paths.
Our contributions establish a new paradigm for theorem proving through verifier-integrated long chain-of-thought (CoT) reasoning, 
and provide open access to our framework and resources for the research community. This work represents a significant step toward making formal reasoning capabilities more accessible and reliable in language models.

\noindent \textbf{Limitations.} As shown in the experimental results, we demonstrate improvements on two SOTA prover models: Kimina-Prover-Preview-Distill-7B and DeepSeek-Prover-V2-7B. 
Through in-depth analysis, we find that when the difficulty of formal statements increases to a certain level, the potential number of active prompts for RL training decreases for two reasons: first, the original problem set for harder levels (such as IMO) is much smaller, second we filter out problems through pass rate (all/none solved) during training. Therefore, for very strong base models (such as DeepSeek-Prover-V2-7B), our RL framework may suffer from limited active training prompts used during training, making our benefit shows a diminishing trend.

\noindent \textbf{Future Work.} Up to this work, we have completed two exploratory studies on how to improve models' cognitive behaviors, including our previous work \texttt{Leanabell-Prover-V1}~\citep{zhang2025leanabell}, especially their reflection and correction capabilities. This has made us realize that there are still the following very important problems that urgently need to be solved:
\begin{itemize}
    \item \textit{Subgoal decomposition via RL}. DeepSeek-Prover-V2~\citep{ren2025deepseek} has proven that subgoal decomposition is an efficient method for solving complex formal problems and scaling synthetic data generation. We conducted experiments with prompt-based approaches for subgoal decomposition but observed consistently low success rates. RL-based methods potentially offer more promising avenues for achieving better results. However, how to effectively leverage RL methods to enhance subgoal decomposition efficiency remains an open research question.
    \item \textit{Partial rollout}. %
    As to address the limitations discussed above, for solved none filtered problems, we can try to concatenate some hints into the prompt such as decomposed subgoals and partial proofs, and perform rollouts on the remaining reasoning and proof parts. A few partial rollout strategies~\citep{bhattacharya2020reinforcement,team2025kimi,beck2025sfo,xu2025not} have demonstrated great potentials in boosting RL training performance for difficult problems.
    \item \textit{Fine-grained rewards}. Based on the detailed verification feedback obtained, in theory, we can construct a structured reward signal. As shown in the Appendix~\ref{app:rewards-design}, we have investigated such methods, however, by far we have not yet obtained favorable results. Yet, we believe that this direction still warrants further exploration.
    \item \textit{Context pruning strategy}. In our verifier-integrated reasoning iteration cycles, we simply concatenate the history of long CoTs and feedback errors in a straightforward manner. By organizing and analyzing the thinking process, generated proofs, and validation feedback, and incorporating the useful information as context into the next iteration cycle, we can reduce context length, improve inference efficiency, and strengthen the model's reflective performance.
\end{itemize}

\setcitestyle{numbers}
\bibliography{main}

\begin{thebibliography}{39}
\providecommand{\natexlab}[1]{#1}
\providecommand{\url}[1]{\texttt{#1}}
\expandafter\ifx\csname urlstyle\endcsname\relax
  \providecommand{\doi}[1]{doi: #1}\else
  \providecommand{\doi}{doi: \begingroup \urlstyle{rm}\Url}\fi

\bibitem[Anthropic(2025)]{claude2025sonnet}
Anthropic.
\newblock Claude 3.7 {S}onnet {S}ystem card.
\newblock 2025.
\newblock URL \url{https://www.anthropic.com/news/claude-3-7-sonnet}.

\bibitem[Azerbayev et~al.(2023)Azerbayev, Piotrowski, Schoelkopf, Ayers, Radev, and Avigad]{azerbayev2023proofnet}
Z.~Azerbayev, B.~Piotrowski, H.~Schoelkopf, E.~W. Ayers, D.~Radev, and J.~Avigad.
\newblock Proofnet: Autoformalizing and formally proving undergraduate-level mathematics.
\newblock \emph{arXiv preprint arXiv:2302.12433}, 2023.

\bibitem[Beck(2025)]{beck2025sfo}
J.~Beck.
\newblock Offline rlaif: Piloting vlm feedback for rl via sfo.
\newblock \emph{arXiv preprint arXiv:2503.01062}, 2025.

\bibitem[Bhattacharya et~al.(2020)Bhattacharya, Badyal, Wheeler, Gil, and Bertsekas]{bhattacharya2020reinforcement}
S.~Bhattacharya, S.~Badyal, T.~Wheeler, S.~Gil, and D.~Bertsekas.
\newblock Reinforcement learning for pomdp: Partitioned rollout and policy iteration with application to autonomous sequential repair problems.
\newblock \emph{IEEE Robotics and Automation Letters}, 5\penalty0 (3):\penalty0 3967--3974, 2020.

\bibitem[Coq(1996)]{coq1996coq}
P.~Coq.
\newblock The coq proof assistant-reference manual.
\newblock \emph{INRIA Rocquencourt and ENS Lyon, version}, 5, 1996.

\bibitem[De~Moura et~al.(2015)De~Moura, Kong, Avigad, Van~Doorn, and von Raumer]{de2015lean}
L.~De~Moura, S.~Kong, J.~Avigad, F.~Van~Doorn, and J.~von Raumer.
\newblock The {L}ean theorem prover (system description).
\newblock In \emph{International Conference on Automated Deduction (CAD)}, 2015.

\bibitem[Dong et~al.(2024)Dong, Lu, Li, Xia, Yu, Zhou, and Zhou]{dong2024self}
G.~Dong, K.~Lu, C.~Li, T.~Xia, B.~Yu, C.~Zhou, and J.~Zhou.
\newblock Self-play with execution feedback: Improving instruction-following capabilities of large language models.
\newblock \emph{arXiv preprint arXiv:2406.13542}, 2024.

\bibitem[Dong and Ma(2025)]{dong2025stp}
K.~Dong and T.~Ma.
\newblock Stp: Self-play llm theorem provers with iterative conjecturing and proving.
\newblock \emph{arXiv preprint arXiv:2502.00212}, 2025.

\bibitem[Feng et~al.(2025)Feng, Huang, Qu, Zhang, Qin, Zhong, Jiang, Chi, and Zhong]{feng2025retool}
J.~Feng, S.~Huang, X.~Qu, G.~Zhang, Y.~Qin, B.~Zhong, C.~Jiang, J.~Chi, and W.~Zhong.
\newblock Retool: Reinforcement learning for strategic tool use in llms.
\newblock \emph{arXiv preprint arXiv:2504.11536}, 2025.

\bibitem[Guo et~al.(2025)Guo, Yang, Zhang, Song, Zhang, Xu, Zhu, Ma, Wang, Bi, et~al.]{guo2025deepseek}
D.~Guo, D.~Yang, H.~Zhang, J.~Song, R.~Zhang, R.~Xu, Q.~Zhu, S.~Ma, P.~Wang, X.~Bi, et~al.
\newblock Deepseek-{R}1: Incentivizing reasoning capability in llms via reinforcement learning.
\newblock \emph{arXiv preprint arXiv:2501.12948}, 2025.

\bibitem[Huth and Ryan(2004)]{huth2004logic}
M.~Huth and M.~Ryan.
\newblock \emph{Logic in Computer Science: Modelling and reasoning about systems}.
\newblock Cambridge university press, 2004.

\bibitem[Jin et~al.(2025)Jin, Zeng, Yue, Yoon, Arik, Wang, Zamani, and Han]{jin2025search}
B.~Jin, H.~Zeng, Z.~Yue, J.~Yoon, S.~Arik, D.~Wang, H.~Zamani, and J.~Han.
\newblock Search-r1: Training llms to reason and leverage search engines with reinforcement learning.
\newblock \emph{arXiv preprint arXiv:2503.09516}, 2025.

\bibitem[Li et~al.(2024)Li, Beeching, Tunstall, Lipkin, Soletskyi, Huang, Rasul, Yu, Jiang, Shen, Qin, Dong, Zhou, Fleureau, Lample, and Polu]{numina_math_datasets}
J.~Li, E.~Beeching, L.~Tunstall, B.~Lipkin, R.~Soletskyi, S.~C. Huang, K.~Rasul, L.~Yu, A.~Jiang, Z.~Shen, Z.~Qin, B.~Dong, L.~Zhou, Y.~Fleureau, G.~Lample, and S.~Polu.
\newblock Numinamath, 2024.

\bibitem[Lin et~al.(2025)Lin, Tang, Lyu, Wu, Lin, Yang, Li, Xia, Chen, Arora, et~al.]{lin2025goedel}
Y.~Lin, S.~Tang, B.~Lyu, J.~Wu, H.~Lin, K.~Yang, J.~Li, M.~Xia, D.~Chen, S.~Arora, et~al.
\newblock Goedel-prover: A frontier model for open-source automated theorem proving.
\newblock \emph{arXiv preprint arXiv:2502.07640}, 2025.

\bibitem[Moura and Ullrich(2021)]{moura2021lean}
L.~d. Moura and S.~Ullrich.
\newblock The lean 4 theorem prover and programming language.
\newblock In \emph{International Conference on Automated Deduction}, 2021.

\bibitem[Nipkow et~al.(2002)Nipkow, Wenzel, and Paulson]{nipkow2002isabelle}
T.~Nipkow, M.~Wenzel, and L.~C. Paulson.
\newblock \emph{Isabelle/HOL: a proof assistant for higher-order logic}.
\newblock Springer, 2002.

\bibitem[Parisi et~al.(2022)Parisi, Zhao, and Fiedel]{parisi2022talm}
A.~Parisi, Y.~Zhao, and N.~Fiedel.
\newblock Talm: Tool augmented language models.
\newblock \emph{arXiv preprint arXiv:2205.12255}, 2022.

\bibitem[Patil et~al.(2024)Patil, Zhang, Wang, and Gonzalez]{patil2024gorilla}
S.~G. Patil, T.~Zhang, X.~Wang, and J.~E. Gonzalez.
\newblock Gorilla: Large language model connected with massive apis.
\newblock \emph{Advances in Neural Information Processing Systems (NeurIPS)}, 2024.

\bibitem[Qian et~al.(2025)Qian, Acikgoz, He, Wang, Chen, Hakkani-T{\"u}r, Tur, and Ji]{qian2025toolrl}
C.~Qian, E.~C. Acikgoz, Q.~He, H.~Wang, X.~Chen, D.~Hakkani-T{\"u}r, G.~Tur, and H.~Ji.
\newblock Toolrl: Reward is all tool learning needs.
\newblock \emph{arXiv preprint arXiv:2504.13958}, 2025.

\bibitem[Ren et~al.(2025)Ren, Shao, Song, Xin, Wang, Zhao, Zhang, Fu, Zhu, Yang, et~al.]{ren2025deepseek}
Z.~Ren, Z.~Shao, J.~Song, H.~Xin, H.~Wang, W.~Zhao, L.~Zhang, Z.~Fu, Q.~Zhu, D.~Yang, et~al.
\newblock Deepseek-prover-v2: Advancing formal mathematical reasoning via reinforcement learning for subgoal decomposition.
\newblock \emph{arXiv preprint arXiv:2504.21801}, 2025.

\bibitem[Schick et~al.(2023)Schick, Dwivedi-Yu, Dess{\`\i}, Raileanu, Lomeli, Hambro, Zettlemoyer, Cancedda, and Scialom]{schick2023toolformer}
T.~Schick, J.~Dwivedi-Yu, R.~Dess{\`\i}, R.~Raileanu, M.~Lomeli, E.~Hambro, L.~Zettlemoyer, N.~Cancedda, and T.~Scialom.
\newblock Toolformer: Language models can teach themselves to use tools.
\newblock \emph{Advances in Neural Information Processing Systems (NeurIPS)}, 2023.

\bibitem[Shao et~al.(2024)Shao, Wang, Zhu, Xu, Song, Bi, Zhang, Zhang, Li, Wu, et~al.]{shao2024deepseekmath}
Z.~Shao, P.~Wang, Q.~Zhu, R.~Xu, J.~Song, X.~Bi, H.~Zhang, M.~Zhang, Y.~Li, Y.~Wu, et~al.
\newblock Deepseekmath: Pushing the limits of mathematical reasoning in open language models.
\newblock \emph{arXiv preprint arXiv:2402.03300}, 2024.

\bibitem[Shinn et~al.(2023)Shinn, Cassano, Gopinath, Narasimhan, and Yao]{shinn2023reflexion}
N.~Shinn, F.~Cassano, A.~Gopinath, K.~Narasimhan, and S.~Yao.
\newblock Reflexion: Language agents with verbal reinforcement learning.
\newblock \emph{Advances in Neural Information Processing Systems (NeurIPS)}, 2023.

\bibitem[Song et~al.(2025)Song, Jiang, Min, Chen, Chen, Zhao, Fang, and Wen]{song2025r1}
H.~Song, J.~Jiang, Y.~Min, J.~Chen, Z.~Chen, W.~X. Zhao, L.~Fang, and J.-R. Wen.
\newblock R1-searcher: Incentivizing the search capability in llms via reinforcement learning.
\newblock \emph{arXiv preprint arXiv:2503.05592}, 2025.

\bibitem[Team et~al.(2025)Team, Du, Gao, Xing, Jiang, Chen, Li, Xiao, Du, Liao, et~al.]{team2025kimi}
K.~Team, A.~Du, B.~Gao, B.~Xing, C.~Jiang, C.~Chen, C.~Li, C.~Xiao, C.~Du, C.~Liao, et~al.
\newblock Kimi k1.5: Scaling reinforcement learning with llms.
\newblock \emph{arXiv preprint arXiv:2501.12599}, 2025.

\bibitem[Team(2024)]{qwen2.5}
Q.~Team.
\newblock Qwen2.5: A party of foundation models, September 2024.
\newblock URL \url{https://qwenlm.github.io/blog/qwen2.5/}.

\bibitem[Team(2025)]{qwq32b}
Q.~Team.
\newblock Qwq-32b: Embracing the power of reinforcement learning, March 2025.
\newblock URL \url{https://qwenlm.github.io/blog/qwq-32b/}.

\bibitem[Tsoukalas et~al.(2024)Tsoukalas, Lee, Jennings, Xin, Ding, Jennings, Thakur, and Chaudhuri]{tsoukalas2024putnambench}
G.~Tsoukalas, J.~Lee, J.~Jennings, J.~Xin, M.~Ding, M.~Jennings, A.~Thakur, and S.~Chaudhuri.
\newblock Putnambench: Evaluating neural theorem-provers on the putnam mathematical competition.
\newblock \emph{arXiv preprint arXiv:2407.11214}, 2024.

\bibitem[Wang et~al.(2025{\natexlab{a}})Wang, Qian, Zhong, Chen, Qiu, Huang, Jin, Wang, Wong, and Ji]{wang2025otc}
H.~Wang, C.~Qian, W.~Zhong, X.~Chen, J.~Qiu, S.~Huang, B.~Jin, M.~Wang, K.-F. Wong, and H.~Ji.
\newblock Otc: Optimal tool calls via reinforcement learning.
\newblock \emph{arXiv preprint arXiv:2504.14870}, 2025{\natexlab{a}}.

\bibitem[Wang et~al.(2025{\natexlab{b}})Wang, Unsal, Lin, Baksys, Liu, Santos, Sung, Vinyes, Ying, Zhu, et~al.]{wang2025kimina}
H.~Wang, M.~Unsal, X.~Lin, M.~Baksys, J.~Liu, M.~D. Santos, F.~Sung, M.~Vinyes, Z.~Ying, Z.~Zhu, et~al.
\newblock Kimina-prover preview: Towards large formal reasoning models with reinforcement learning.
\newblock \emph{arXiv preprint arXiv:2504.11354}, 2025{\natexlab{b}}.

\bibitem[Wang et~al.(2024)Wang, Zhang, Jia, Pan, Diao, Pi, and Zhang]{wang2024theoremllama}
R.~Wang, J.~Zhang, Y.~Jia, R.~Pan, S.~Diao, R.~Pi, and T.~Zhang.
\newblock Theoremllama: Transforming general-purpose llms into lean4 experts.
\newblock \emph{arXiv preprint arXiv:2407.03203}, 2024.

\bibitem[Xin et~al.(2024{\natexlab{a}})Xin, Guo, Shao, Ren, Zhu, Liu, Ruan, Li, and Liang]{xin2024deepseek-v1}
H.~Xin, D.~Guo, Z.~Shao, Z.~Ren, Q.~Zhu, B.~Liu, C.~Ruan, W.~Li, and X.~Liang.
\newblock Deepseek-prover: Advancing theorem proving in llms through large-scale synthetic data.
\newblock \emph{arXiv preprint arXiv:2405.14333}, 2024{\natexlab{a}}.

\bibitem[Xin et~al.(2024{\natexlab{b}})Xin, Ren, Song, Shao, Zhao, Wang, Liu, Zhang, Lu, Du, et~al.]{xin2024deepseek-v1.5}
H.~Xin, Z.~Ren, J.~Song, Z.~Shao, W.~Zhao, H.~Wang, B.~Liu, L.~Zhang, X.~Lu, Q.~Du, et~al.
\newblock Deepseek-prover-v1. 5: Harnessing proof assistant feedback for reinforcement learning and monte-carlo tree search.
\newblock \emph{arXiv preprint arXiv:2408.08152}, 2024{\natexlab{b}}.

\bibitem[Xu et~al.(2025)Xu, Savani, Fang, and Kolter]{xu2025not}
Y.~E. Xu, Y.~Savani, F.~Fang, and Z.~Kolter.
\newblock Not all rollouts are useful: Down-sampling rollouts in llm reinforcement learning.
\newblock \emph{arXiv preprint arXiv:2504.13818}, 2025.

\bibitem[Yao et~al.(2023)Yao, Zhao, Yu, Du, Shafran, Narasimhan, and Cao]{yao2023react}
S.~Yao, J.~Zhao, D.~Yu, N.~Du, I.~Shafran, K.~Narasimhan, and Y.~Cao.
\newblock React: Synergizing reasoning and acting in language models.
\newblock In \emph{International Conference on Learning Representations (ICLR)}, 2023.

\bibitem[Yu et~al.(2025)Yu, Zhang, Zhu, Yuan, Zuo, Yue, Dai, Fan, Liu, Liu, et~al.]{yu2025dapo}
Q.~Yu, Z.~Zhang, R.~Zhu, Y.~Yuan, X.~Zuo, Y.~Yue, W.~Dai, T.~Fan, G.~Liu, L.~Liu, et~al.
\newblock Dapo: An open-source llm reinforcement learning system at scale.
\newblock \emph{arXiv preprint arXiv:2503.14476}, 2025.

\bibitem[Zhang et~al.(2025)Zhang, Wang, Ji, Liu, Yue, Zhang, Zhang, Zhou, and Gai]{zhang2025leanabell}
J.~Zhang, Q.~Wang, X.~Ji, Y.~Liu, Y.~Yue, F.~Zhang, D.~Zhang, G.~Zhou, and K.~Gai.
\newblock Leanabell-prover: Posttraining scaling in formal reasoning.
\newblock \emph{arXiv preprint arXiv:2504.06122}, 2025.

\bibitem[Zheng et~al.(2021)Zheng, Han, and Polu]{zheng2021minif2f}
K.~Zheng, J.~M. Han, and S.~Polu.
\newblock Minif2f: a cross-system benchmark for formal olympiad-level mathematics.
\newblock \emph{arXiv preprint arXiv:2109.00110}, 2021.

\bibitem[Ziegenbein et~al.(2024)Ziegenbein, Skitalinskaya, Makou, and Wachsmuth]{ziegenbein2024llm}
T.~Ziegenbein, G.~Skitalinskaya, A.~B. Makou, and H.~Wachsmuth.
\newblock Llm-based rewriting of inappropriate argumentation using reinforcement learning from machine feedback.
\newblock \emph{arXiv preprint arXiv:2406.03363}, 2024.

\end{thebibliography}
\setcitestyle{authoryear}

\appendix
\appendix

\section{Cold-start Details}
\label{app:cold_start}

As shown in Figure~\ref{fig:correcting-prompt} and Figure~\ref{fig:cold-start-prompt}, we present the prompts for correcting the proof errors and synthesizing data with long CoT with Claude-Sonnet-3.7, respectively.

\begin{figure}[htpb]
    \centering
    \begin{center}
    \begin{promptbox}[Prompt for Correcting the Proof Errors]
    You are an AI assistant proficient in using Lean 4 to solve mathematical problems. At the beginning, when solving a problem, you need to think step by step and generate complete Lean 4 code, which is then compiled in the Lean 4 environment. The Lean 4 compilation is executed in a sandbox environment, and the compilation results are returned. These results may include error logs to assist with your reasoning. Now, continue writing based on the original thinking process that includes the compilation results to generate an analysis and complete, correct Lean 4 code.

    \

    In the original thinking process provided to you, the Lean 4 code block generated by the model is enclosed between <code> and </code>, while the compilation result of the Lean 4 environment is enclosed between <interpreter> and </interpreter>.
    
    \

    GOAL: 
    
    Continue writing based on the original thinking process. Please first analyze the user's question, the original thinking process, the generated Lean 4 code block, and the error logs, and provide the result of the error analysis. Then, present the complete Lean 4 code block that you think is correct. After generating the entire error analysis result, end it with </think>. Then generate the Lean 4 code block, starting with \text{`}\text{`}\text{`}lean4 and ending with \text{'}\text{'}\text{'}.
    
    \

    The output format is as follows:
    
    \{Your analysis result\}</think>

    \text{`}\text{`}\text{`}lean4\newline\{The revised Lean 4 code block\}\newline\text{'}\text{'}\text{'}

    \

    User Question:
    
    \{question\}

    \

    Original Thinking Process:
    
    <original\_thinking\_process> {original\_response} </original\_thinking\_process>
    \end{promptbox}
    \end{center}
    \caption{Prompt for correcting the proof errors through Claude-3.7-Sonnet.}
    \label{fig:correcting-prompt}
\end{figure}

\begin{figure}[ht]
    \centering
    \begin{center}
    \begin{promptbox}[Prompt for Synthesizing Data with Long CoT]
You are an AI assistant proficient in using Lean 4 to solve mathematical problems. At the beginning, when solving a problem, you need to think step by step and generate complete Lean 4 code, which is then compiled in the Lean 4 environment. The Lean 4 compilation is executed in a sandbox environment, and the compilation results are returned. These results may include error logs to assist with your reasoning. Now, continue writing based on the original thinking process that includes the compilation results to generate an analysis and complete, correct Lean 4 code.
\newline

In the original thinking process provided to you, the Lean 4 code block generated by the model is enclosed between \text{`}\text{`}\text{`}lean4 and \text{'}\text{'}\text{'}, while the compilation result of the Lean 4 environment is enclosed between <compiler\_results> and </compiler\_results>.
\newline

GOAL: \newline
Continue writing based on the original thinking process. Please first analyze the user's question, the original thinking process, the generated Lean 4 code block, and the error logs, and provide the result of the error analysis. Then, present the complete Lean 4 code block that you think is correct. After generating the entire error analysis result, end it with </think>. Then generate the Lean 4 code block, starting with \text{`}\text{`}\text{`}lean4 and ending with \text{'}\text{'}\text{'}.\newline

The output format is as follows:\newline
\{\{Your analysis result\}\}\newline\text{`}\text{`}\text{`}lean4{{The revised Lean 4 code block}}\text{'}\text{'}\text{'}\newline

User Question:\newline
\{\texttt{question}\}\newline

Original Thinking Process:\newline
<original\_thinking\_process> \{original\_response\} </original\_thinking\_process>
    \end{promptbox}
    \end{center}
    \caption{Prompt for synthesizing data with long CoT through Claude-3.7-Sonnet.}
    \label{fig:cold-start-prompt}
\end{figure}

\begin{figure}[htpb]
    \centering
    \begin{center}
    Example input (i.e., between \texttt{<code></code>}) Lean 4 code:
    \begin{lstlisting}[frame=single]
import Mathlib 
import Aesop 
set_option maxHeartbeats 0 
open BigOperators Real Nat Topology Rat

/-- After moving the line $y=-3x+5$ down $3$ units, the equation of the resulting line is ______. Show that it is y = -3x + 2.-/
theorem my_favorite_theorem (y : ℝ → ℝ) (h : y = fun x => -3 * x + 5) :
    ∀ x, y x - 3 = -3 * x + 2 := by 
  intro x
  rw [h]
  ring\end{lstlisting}
    Corresponding AST output in JSON format:
    \begin{lstlisting}[frame=single]
"ast": {
    "tactics": [
        {
            "stateBefore": "y : ℝ → ℝ\nh : y = fun x => -3 * x + 5\n⊢ ∀ (x : ℝ), y x - 3 = -3 * x + 2",
            "stateAfter": "y : ℝ → ℝ\nh : y = fun x => -3 * x + 5\nx : ℝ\n⊢ y x - 3 = -3 * x + 2",
            "pos": 342,
            "endPos": 349
        },
        {
            "stateBefore": "y : ℝ → ℝ\nh : y = fun x => -3 * x + 5\nx : ℝ\n⊢ y x - 3 = -3 * x + 2",
            "stateAfter": "y : ℝ → ℝ\nh : y = fun x => -3 * x + 5\nx : ℝ\n⊢ (fun x => -3 * x + 5) x - 3 = -3 * x + 2",
            "pos": 352,
            "endPos": 358
        },
        {
            "stateBefore": "y : ℝ → ℝ\nh : y = fun x => -3 * x + 5\nx : ℝ\n⊢ y x - 3 = -3 * x + 2",
            "stateAfter": "y : ℝ → ℝ\nh : y = fun x => -3 * x + 5\nx : ℝ\n⊢ (fun x => -3 * x + 5) x - 3 = -3 * x + 2",
            "pos": 356,
            "endPos": 357
        },
        {
            "stateBefore": "y : ℝ → ℝ\nh : y = fun x => -3 * x + 5\nx : ℝ\n⊢ (fun x => -3 * x + 5) x - 3 = -3 * x + 2",
            "stateAfter": "no goals",
            "pos": 361,
            "endPos": 364
        }
    ],
    "premises": [0],
    "declarations": [4]
}\end{lstlisting}
    \end{center}
    \vspace{-1.5em}
    \caption{Example of feedback Abstract Syntax Tree (AST) for the generated Lean 4 code.}
    \label{fig:ast-feedback}
\end{figure}

\begin{figure}[htpb]
    \centering
    \begin{center}
    \includegraphics[width=\textwidth]{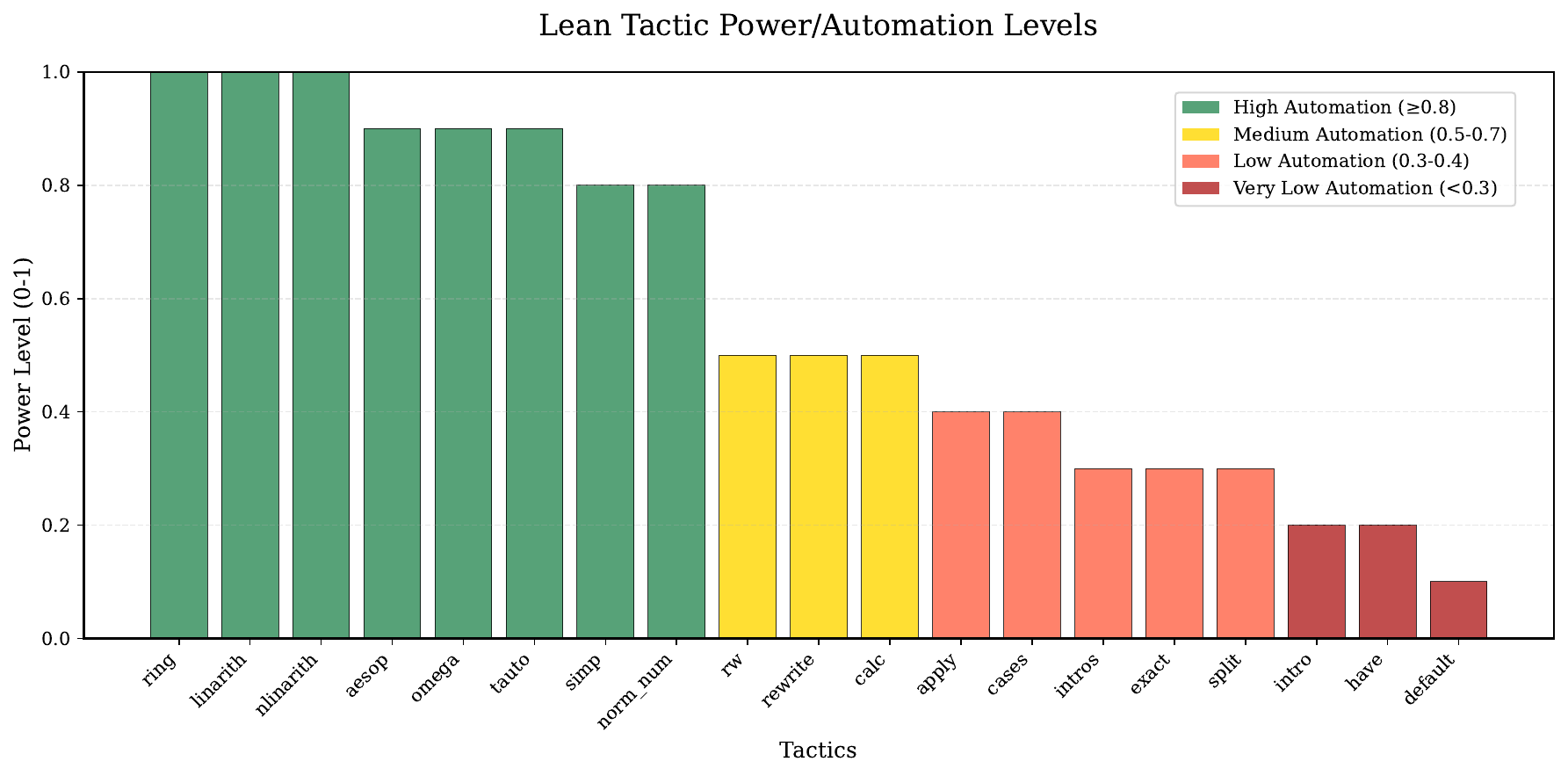}
    \caption{Lean 4 tactic power/automation levels.}
    \label{fig:tactic-complexity}
    \end{center}
\end{figure}

\section{On Reward Design}
\label{app:rewards-design}

Besides the aforementioned validation status rewards $R_\text{failed}$ and $R_\text{success}$, we also explore a structured reward signal in the RL training. Although we do not obtain obvious improvements from such structured reward signal, it can be a useful guidance for researchers who want to explore this direction in the future. Given an input Lean 4 code and its Abstract Syntax Tree (AST) obtained by the Lean 4 verifier, as shown in Figure~\ref{fig:ast-feedback} in Appendix, we analyze to evaluate proof quality. We consider five main aspects for the reward design. 
\begin{itemize}
    \item  \textit{Tactic Count Reward.} Tactic count measures the total number of individual proof steps, where fewer tactics often signal more streamlined and elegant reasoning. Given a formal proof $p$ and a function $Count(\cdot)$ that compute the proof steps, the tactic count reward is formulated as: 
    \begin{equation}
        R_\text{tactic count} = {Count}(p)
    \end{equation} 
    \item \textit{Automation Level Reward.} As shown in the Figure~\ref{fig:tactic-complexity}, we show the different power or automation level of tactics. The power/automation level is associated with the frequency of a tactic. Automation level reward evaluates the sophistication of the proof steps themselves—powerful tactics like \texttt{ring} for algebraic manipulation, \texttt{simp} for simplification, or \texttt{linarith} for linear arithmetic can encapsulate complex mathematical reasoning within a single command, potentially replacing dozens of manual steps. For the $i$-th tactic in the proof, let $v_i$  be its power value from the pre-defined ``tactic power'' dictionary, as shown in Figure~\ref{fig:tactic-complexity}. Thus, we have:
    \begin{equation}
        R_\text{automation} = \frac{1}{n}\sum_{i=1}^n v_i
    \end{equation}
    \item \textit{State Change Efficiency Reward.} It quantifies how much progress each tactic makes toward the proof goal by comparing the proof state before and after each step, helping identify which tactics contribute most meaningfully to the overall proof construction. For the $i$-th tactic in a proof $p$ with $g_i^{\text{before}}$ goals before execution and $g_i^{\text{after}}$ goals after execution, we have the state change efficiency reward:
    \begin{equation}
        R_\text{state change} = \max(0, 1 - \frac{g_i^{\text{after}}}{g_i^{\text{before}}}), \quad \text{for} \quad g_i^{\text{before}} > 0
    \end{equation}
    Notably, $g_i^{\text{after}}=0$ refers to the $i$-th tactic solves all goals, $g_i^{\text{after}}=g_i^{\text{before}}$ refers to no progress made, $g_i^{\text{after}}>g_i^{\text{before}}$ the $i$-th tactic creates more goals, and $0<g_i^{\text{after}}<g_i^{\text{before}}$ means achieving partial progress. 
\end{itemize}

Finally, we obtain the overall reward by combining the multiple rewards:
\begin{equation}
R_\text{final} = \begin{cases}
R_{\text{failed}}  & \quad \text{if failed} \\
R_{\text{success}} + \lambda_{tc}R_{\text{tactic count}} + \lambda_{at}R_{\text{automation}} + \lambda_{sc}R_{\text{state change}} & \quad \text{if success}
\end{cases}
\end{equation}
where $\lambda_{tc}$, $\lambda_{at}$ and $\lambda_{sc}$ are hyperparameters to control the reward strength.

As shown in Table~\ref{tab:reward-ablation}, we employ Kimina-Prover-Preview-Distill-7B as a baseline model and vary the reward used in the RL algorithm. We find that when using $R_\text{tactic count}$, we achieve slight gains, but the other two rewards (\textit{i.e.}, $R_\text{automation}$ and $R_\text{state change}$) both perform slightly worse than using the simple $R_\text{failed}$ and $R_\text{success}$ reward strategy.

\begin{table*}[htp]
\setlength{\tabcolsep}{0.2in}
\begin{center}
\small
\begin{tabular}{lccccc}
\toprule
    \textbf{Method} & \textbf{Sample budgets} & \textbf{MiniF2F-test} \\
    \midrule
    Kimina-Prover-Preview-Distill-7B~\citep{wang2025kimina}  & \multirow{5}{*}{32} & 63.1\% \\
     \quad + Success/failure reward only & & 68.4\% \\
     \quad\quad w/ $\lambda_{tc}=1.0$ & &  69.2\%\\
     \quad\quad w/ $\lambda_{at}=1.0$ & &  68.0\%\\
     \quad\quad w/ $\lambda_{sc}=1.0$ & &  67.2\%\\
    \bottomrule
\end{tabular}
\caption{
Comparison between the simplest success/failure reward and our proposed fine-grained rewards. 
}
\label{tab:reward-ablation} %
\end{center}
\end{table*}

\begin{figure}[htpb]
\begin{minipage}{0.24\textwidth}
    \includegraphics[width=\textwidth]{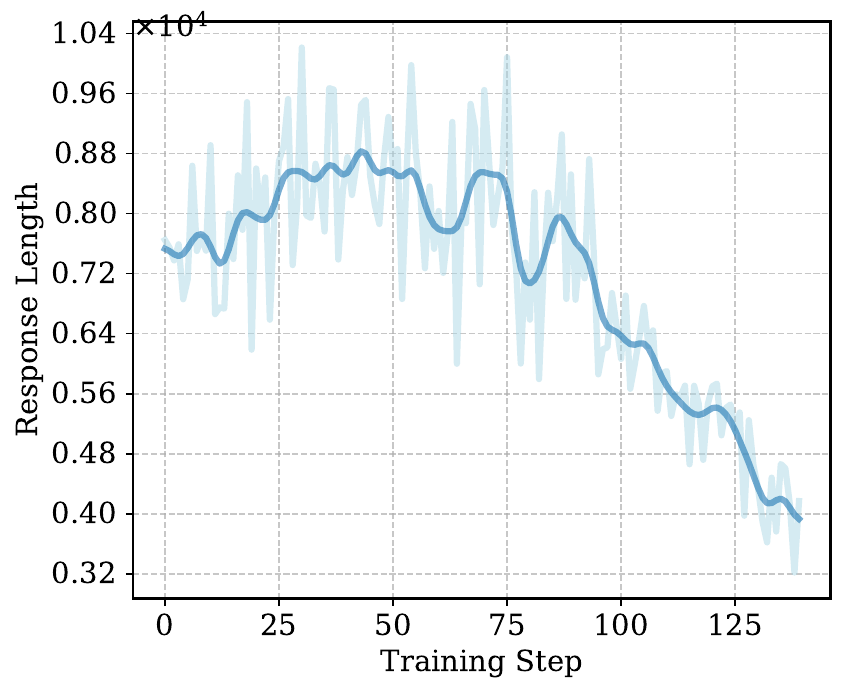}
\end{minipage}
 \hfill
\begin{minipage}{0.24\textwidth}
    \centering
    \includegraphics[width=\textwidth]{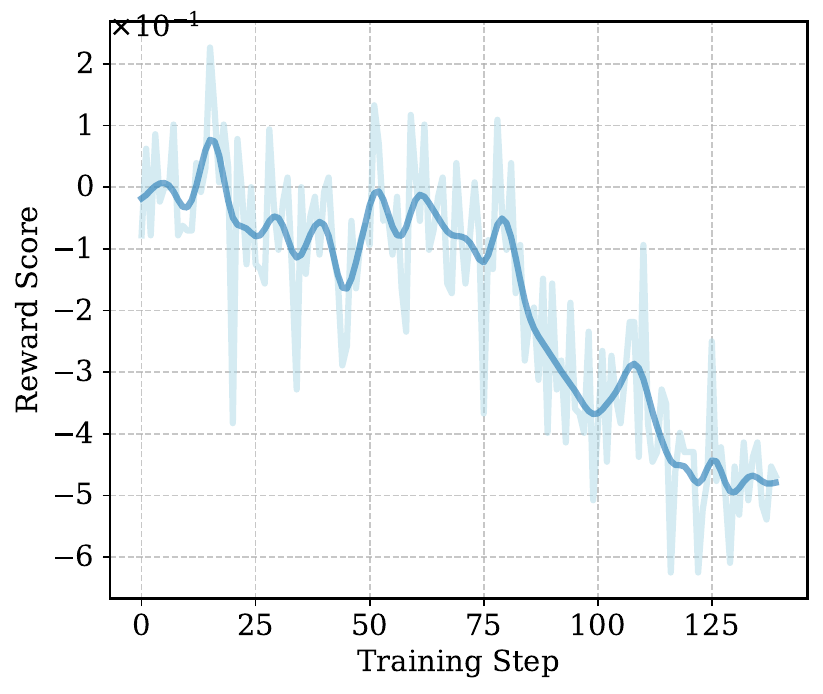}
\end{minipage}
\hfill
\begin{minipage}{0.24\textwidth}
    \centering
    \includegraphics[width=\textwidth]{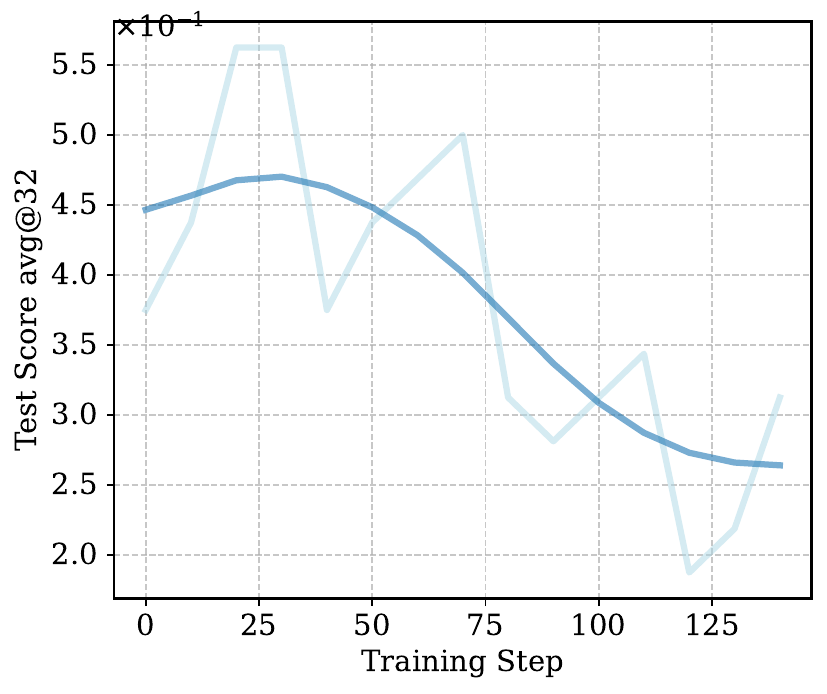}
\end{minipage}
\hfill
\begin{minipage}{0.24\textwidth}
    \centering
    \includegraphics[width=\textwidth]{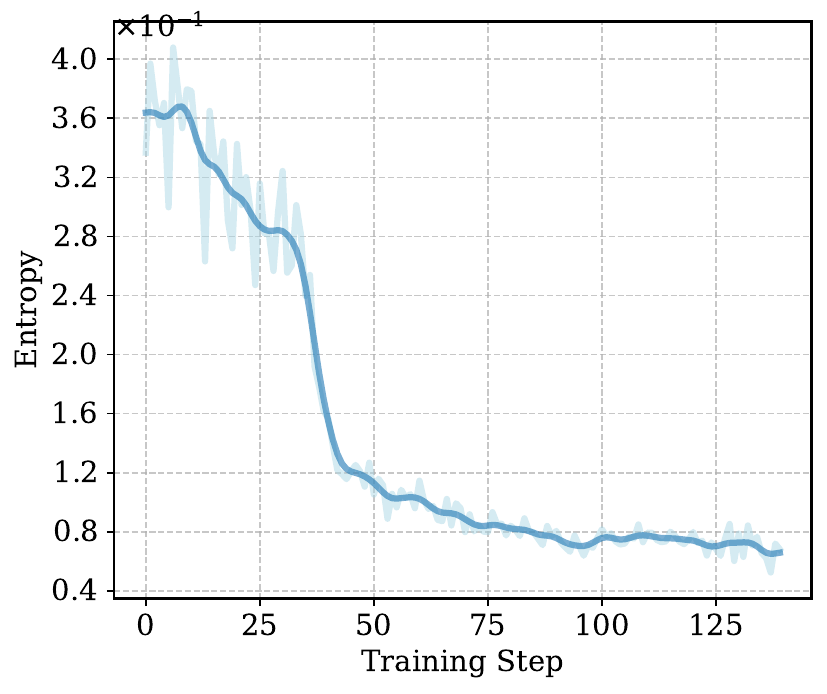}
\end{minipage}
\caption{Several core records in the RL training with GRPO~\citep{shao2024deepseekmath} algorithm and DeepSeek-Prover-V2-7B model, including \textit{Response Length}, \textit{Reward Score}, \textit{Test Score avg@32}, and \textit{Entropy}.}
\label{fig:grpo-records}
\end{figure}

\section{On RL Algorithm}
\label{app:rl-algoritm}

In our main experiments, we follow DAPO~\citep{yu2025dapo} and employ a token-level loss in the RL algorithm. We heve investigated the impact of loss aggregation strategies on model training stability. For comparison, we have explored 
a sample-level loss calculation (similar to GRPO~\citep{shao2024deepseekmath}), which involves first averaging the losses by token within each sample and then aggregating the losses across samples. However, this approach proves problematic for this verifier-integrated reasoning task, where responses vary significantly in length and are generally longer than in conventional reasoning scenarios. The averaging process effectively reduces the learning signal from longer responses, as each token in these responses contributes less to the overall loss. This reduction impairs the model's ability to learn from the rich reasoning patterns present in the long CoTs.
As shown in Figure~\ref{fig:grpo-records}, we encounter training collapse problems with such a sample-level loss. 
In contrast, the token-level loss computation employed by DAPO maintains uniform gradient contributions across all tokens regardless of sequence length, thereby resulting in more stable and effective training for our task.

\section{Success Example}
\label{app:success_example}

We show an example that successfully correct the errors with our proposed verifier-integrated in long CoT reasoning. As shown in Figure~\ref{fig:success_example-1}and Figure~\ref{fig:success_example-2}, we take our Leanabell-Prover-V2-KM as an example, where there is an error about ``\texttt{type mismatch}'' in validation feedback at the first round reasoning. After being informed of such errors, the model generated the correct proof through reflective attempts.  Note that due to differences in the base models themselves, there are some differences bewteen our Leanabell-Prover-V2-KM and Leanabell-Prover-V2-DS in reasoning style. However, this does not affect the overall reasoning process, so additional examples will not be provided here for demonstration.

\begin{figure}[p]
    \centering
    \begin{center}
    Example prompt for Lean 4 code completion:
    \begin{lstlisting}[frame=single]
You are an expert in mathematics and Lean 4. ... 
Think step by step and solve the following problem in Lean 4. ...

# Problem: If $r$ is rational $(r \neq 0)$ and $x$ is irrational, prove that $rx$ is irrational.
# Formal statement: 
```lean4
...
theorem exercise_1_1b
(x : ℝ)
(y : ℚ)
(h : y ≠ 0)
: ( Irrational x ) -> Irrational ( x * y ) := by
'''\end{lstlisting}
    Output long CoT, Lean 4 code and validation feedback:
    \begin{lstlisting}[frame=single]
<think>
# Proving that the Product of a Non-Zero Rational Number and an Irrational Number is Irrational
To prove that if $r$ is a non-zero rational number and $x$ is irrational, then $rx$ is irrational, we can use the properties of rational and irrational numbers. 
First, let's think about why this is true. If $r$ is rational and non-zero, and $x$ is irrational, then $rx$ cannot be rational.
...
<code>
```lean4
...
/-- If $r$ is rational $(r \neq 0)$ and $x$ is irrational, prove that $rx$ is irrational.-/
theorem exercise_1_1b
(x : ℝ)
(y : ℚ)
(h : y ≠ 0)
: ( Irrational x ) -> Irrational ( x * y ) := by
  intro hx
  have hy : (y : ℝ) ≠ 0 := by exact_mod_cast h
  have h1 : Irrational (x * y) := by
    simpa using hx.mul_rat hy
  exact h1
'''
</code>
<interpreter>
Compilation failed.\nFind an error at line 24\napplication type mismatch\n  Irrational.mul_rat hx hy\nargument\n  hy\nhas type\n  ↑y ≠ 0 : Prop\nbut is expected to have type\n  y ≠ 0 : Prop
</interpreter>\end{lstlisting}
    \end{center}
    \vspace{-1.5em}
    \caption{Example of correcting the errors with our proposed verifier-integrated in long CoT reasoning (Part I).}
    \label{fig:success_example-1}
\end{figure}

\begin{figure}[p]
    \centering
    \begin{center}
    \vspace{-2em}
    Corrected proof and final validation feedback:
    \begin{lstlisting}[frame=single]
It seems that the Lean 4 compilation failed and returned some error logs. Let me check and analyze the error logs.\n\nThe error in the original code is related to the type mismatch in the `hx.mul_rat hy` expression. The error message indicates that `hy` has type `↑y ≠ 0 : Prop` but is expected to have type `y ≠ 0 : Prop`.

The issue is that we need to cast `hy` from `(y : ℝ) ≠ 0` back to `y ≠ 0`. The hypothesis `h` is already `y ≠ 0` in the context (where `y` is of type `ℚ`), but we're trying to use it with `hx.mul_rat hy` where `hy` is cast to `(y : ℝ) ≠ 0`.

To fix this, we need to explicitly cast `hy` back to the original type `y ≠ 0`. We can do this using the `exact_mod_cast` tactic to convert `hy` back to the expected type.
</think>
<code>
```lean4
theorem exercise_1_1b
(x : ℝ)
(y : ℚ)
(h : y ≠ 0)
: ( Irrational x ) -> Irrational ( x * y ) := by
  intro hx
  have hy : y ≠ 0 := h
  have h1 : Irrational (x * y) := by
    simpa using hx.mul_rat hy
  exact h1
'''
</code>
<interpreter>
Compilation Success!
</interpreter>\end{lstlisting}
    \end{center}
    \caption{Example of correcting the errors with our proposed verifier-integrated in long CoT reasoning (Part II).}
    \label{fig:success_example-2}
\end{figure}

\end{CJK*}
\end{document}